\documentclass{article}

\usepackage[utf8]{inputenc}
\usepackage[dvipsnames]{xcolor}
\usepackage[frozencache=true,cachedir=minted-cache]{minted}
\usepackage{xcolor}
\usepackage{bm}
\usepackage{physics}
\usepackage{multirow}
\usepackage{pifont}
\usepackage{enumitem}
\usepackage{array}
\newcolumntype{P}[1]{>{\centering\arraybackslash}p{#1}}

\definecolor{LightGray}{gray}{0.98}
\newcommand{\codefig}[1]{
    \inputminted[
        fontsize=\scriptsize, bgcolor=LightGray
    ]{python}{code/#1}
    \vspace{-0.6cm}
}

\usepackage{microtype}
\usepackage{graphicx}
\usepackage{subfigure}
\usepackage{booktabs}
\usepackage{hyperref}

\usepackage[accepted]{icml2023}

\usepackage{amsmath}
\usepackage{amssymb}
\usepackage{mathtools}
\usepackage{amsthm}
\usepackage[capitalize,noabbrev]{cleveref}

\theoremstyle{plain}
\newtheorem{theorem}{Theorem}[section]
\newtheorem{proposition}[theorem]{Proposition}

\theoremstyle{definition}

\theoremstyle{remark}

\icmltitlerunning{Unit Scaling}

\begin{document}

\twocolumn[
\icmltitle{
Unit Scaling: Out-of-the-Box Low-Precision Training
}
\begin{icmlauthorlist}
\icmlauthor{Charlie Blake}{gc}
\icmlauthor{Douglas Orr}{gc}
\icmlauthor{Carlo Luschi}{gc}
\end{icmlauthorlist}
\icmlaffiliation{gc}{Graphcore Research, United Kingdom}
\icmlcorrespondingauthor{Charlie Blake}{charlieb@graphcore.ai}
\icmlcorrespondingauthor{Douglas Orr}{douglaso@graphcore.ai}

\icmlkeywords{Machine Learning, ICML}
\vskip 0.3in
]
\printAffiliationsAndNotice{}


    \section{{Abstract}}
    We present {\em unit scaling}, a paradigm for designing deep learning models that simplifies the use of low-precision number formats.
Training in FP16 or the recently proposed FP8 formats offers substantial efficiency gains, but can lack sufficient range for out-of-the-box training.
Unit scaling addresses this by introducing a principled approach to model numerics: seeking unit variance of all weights, activations and gradients at initialisation.
Unlike alternative methods, this approach neither requires multiple training runs to find a suitable scale nor has significant computational overhead.
We demonstrate the efficacy of unit scaling across a range of models and optimisers.
We further show that existing models can be adapted to be unit-scaled, training BERT\textsubscript{LARGE} in FP16 and then FP8 with no degradation in accuracy.%

    \section{{Introduction}}
    The development of algorithms that efficiently leverage available hardware has been key to the substantial advances seen in deep learning over the last decade \citep{Sutton19, Hooker21}.

With the increase in size of state-of-the-art models, hardware-efficiency is also motivated by the need to lower the costs of training. These have grown to become substantial---in terms of money, time, and environmental impact \citep{Strubell19, Chowdhery22, Luccioni22}.

However, with the end of Moore's law and Dennard scaling \citep{Esmaeilzadeh11, Theis17}, increased transistor density can no longer be relied upon to provide a simple path towards greater efficiency, and other techniques must be leveraged. One such technique is the use of low-precision number formats. The gains to be had here are considerable: compute, memory and bandwidth usage all depend on the bit-width of a format.

\begin{figure}[t]
    \vspace{-0.3cm}
    \begin{center}
    \includegraphics[width=0.5\textwidth, trim = 0.1cm 0.2cm 0.4cm 0.2cm, clip]{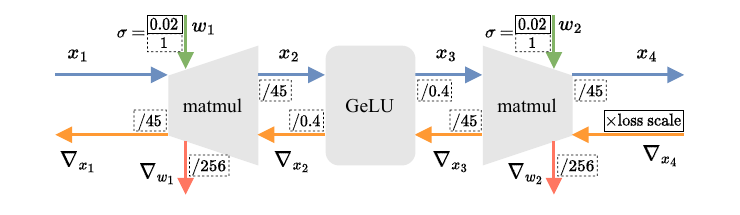}
    \includegraphics[width=0.5\textwidth, trim = 2.5cm 2.5cm 2.5cm 2.5cm, clip]{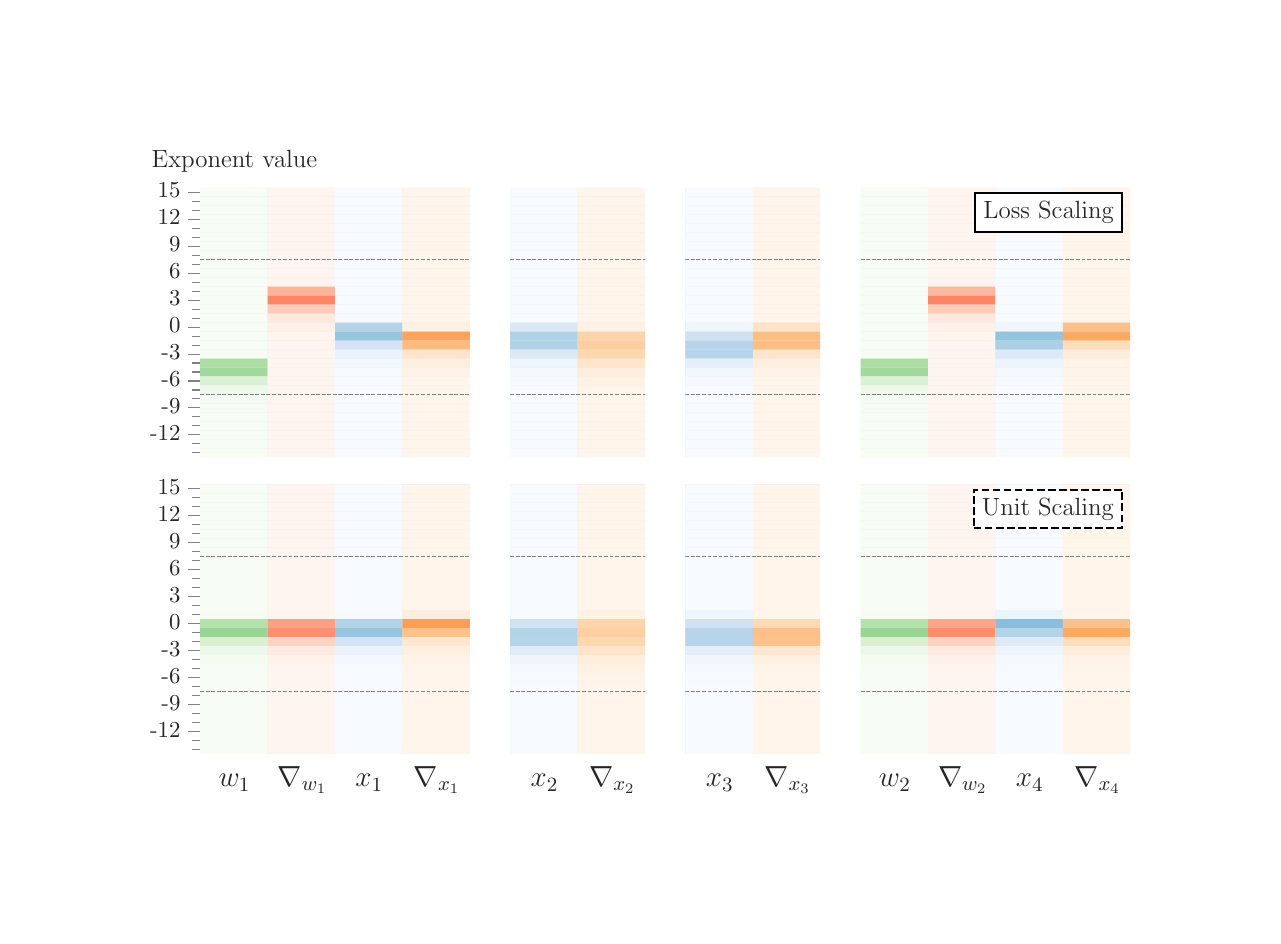}
    \end{center}
    \vspace{-0.3cm}
    \caption{
    \textit{Above:} Unit scaling of an FFN layer. We multiply each tensor by a fixed scalar to achieve consistent scale, no longer requiring a loss scale to control the scale of $\nabla_{x_4}$. Hyperparameters here are the same as those in our BERT\textsubscript{LARGE} experiments (Table~\ref{tab:mask_hyperparameters}).
    \vspace{0.5em}\\\hspace{\textwidth}
    \textit{Below}: A histogram of exponent values at initialisation for the above FFN, with shade indicating bin density.
    The $y$-axis reflects exponent values available in FP16, while dashed  lines show the max/min exponents of the FP8~E4 format of \citet{Noune22}. 
    }
    \label{fig:illustration}
    \vspace{-0.7cm}
\end{figure}

Unlike inference, where integer quantisation is possible \citep{Jacob18}, for training, floating point formats are required~\citep{Noune22, Micikevicius22, Kuzmin22}. The traditional approach of using 32-bit floats is being superseded by mixed precision strategies, which place many values into 16-bit formats~\citep{Micikevicius18}. Furthermore, 8-bit floating-point hardware is becoming available \citep{Graphcore22, Nvidia22}, with the potential for accurate 8-bit training already demonstrated~\citep{Wang18, Sun19, Noune22, Micikevicius22}.

However, the use of low-precision formats introduces new difficulties, reducing the absolute range of representable values and increasing quantisation noise. Existing techniques to address these issues either introduce additional overhead or require manual tuning. An approach is needed which is both accurate and places minimal burden on the user.

To this end, we present {\em unit scaling}: a technique for model design that operates on the principle of ideal scaling at initialisation (unit variance for activations, weights and gradients). This is achieved by considering how each operation in the model affects the variance of different tensors, and introducing fixed scaling factors to counteract changes.

Empirically, we show that unit scaling aligns values much closer to the centre of the representable range than conventional loss scaling \citep{Micikevicius18}, and removes the need for a scaling hyperparameter to be swept. None of our experiments require dynamic re-scaling of values, indicating robustness to shifting distributions during training.

\subsection{Contributions}
In this paper we make the following contributions:
\begin{enumerate}[itemsep=-0.2em]
\vspace{-0.7em}
\item We provide an analysis of how scale changes as a result of operations within a typical model, and the challenges this introduces for low-precision training.
\item We present unit scaling: a method for combating changes in scale, along with an implementation recipe and code examples.
\item We validate unit scaling empirically across a range of models and optimisers.
\item For the first time, we show training of BERT\textsubscript{BASE} and BERT\textsubscript{LARGE} in FP16 without loss scaling. We then go a step further, training successfully in FP8, still without degradation.
\vspace{-0.7em}
\end{enumerate}

We emphasise that our method works out-of-the-box, with no extra sweeps or hyperparameters, demonstrating the effectiveness of unit scaling for simplifying the use of low-precision formats.

    \section{{Background}}
    \begin{figure}
\centering
\includegraphics[width=0.45\textwidth]{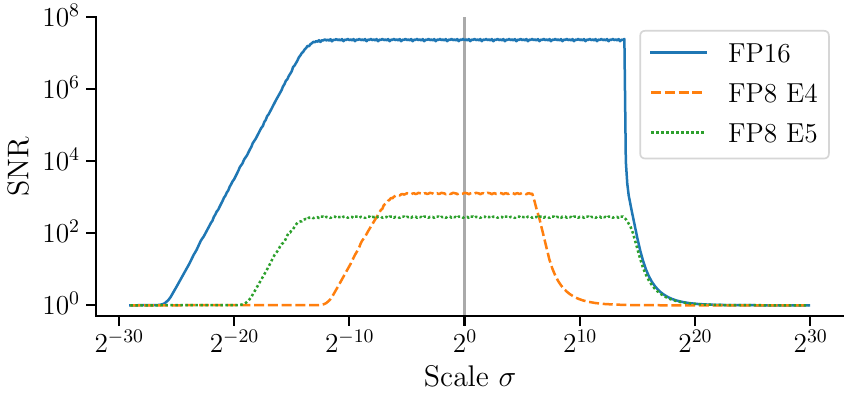}
\vspace{-0.9em}
\caption{The signal to noise ratio (SNR) of samples from a normal distribution, quantised in FP16 and FP8, as a function of the distribution's scale.}
\label{fig:signal_to_noise}
\vspace{-0.9em}
\end{figure}

\begin{table*}
    \caption{A comparison of techniques for low precision training. `$\sim$' indicates that this method ideally requires no tuning, but in practice may introduce hyperparameters that need to be swept.}
    \label{tab:method_comparison}
    \centering
    \vspace{0.6em}
    \begin{tabular}{lP{3.0cm}P{2.8cm}P{3.3cm}P{1.6cm}}\toprule
        Method & Fine-grained scaling & No tuning required & Adapts during training \\
        \midrule
        Loss scaling & × & × & ×\\
        Automatic loss scaling & × & \checkmark & \checkmark \\
        Automatic per-tensor scaling & \checkmark & $\sim$ & \checkmark\\
        Unit scaling & \checkmark & \checkmark & ×\\
        \bottomrule
    \end{tabular}
\end{table*}

\subsection{Floating-point formats for deep learning}
\label{sec:formats}

\paragraph{Definition}

The conventional representation used for floating point numbers is defined by the IEEE 754 standard \citep{IEEE2019}. In this standard, a binary floating point format can be defined by specifying the number of exponent bits, $E$, and the number of mantissa bits, $M$. A value within such a format is defined by a sign bit, exponent and mantissa value. Each is represented using a bit-string of the requisite length (with values $b_{\textrm{sign}}, b_{\textrm{exp}}, b_{\textrm{mant}}$ respectively), which are interpreted as follows:
\[\begin{aligned}
\textrm{exponent} &= b_{\textrm{exp}} - \textrm{bias},\quad \; (\textrm{bias}=2^{E-1}-1)
\\
\textrm{mantissa} &= 1 + \frac{b_{\textrm{mant}}}{2^M},
\\
\textrm{value}&=(-1)^{b_{\textrm{sign}}} \times 2^{\textrm{exponent}}\times \textrm{mantissa}
\end{aligned}\]
There are also a small number of `special values' which denote bit-strings to which the above interpretation does not apply. These represent infinities, NaN (not-a-number) and a range of `subnormal numbers' which allow for the representation of even smaller (absolute) values.

Common floating point formats used in machine learning that implement the IEEE 754 standard are shown in Table~\ref{tab:fp_formats}. The term \textit{low precision} typically refers to all formats requiring fewer than 32 bits. More recently, two kinds of FP8 format have been proposed, which we term E4 and E5, i.e. $(E,M)=(4,3) \textrm{ or } (5,2)$. These are similar to the IEEE 754 standard, but contain differences, especially for the representation of special values. These formats are covered in detail in Appendix~\ref{app:fp8_formats}.

\paragraph{Quantisation error}

Formats with more exponent bits are able to represent a wider range of values, whereas those with more mantissa bits have smaller gaps between represented values.
This trade-off between range and precision can be framed in terms of \textit{quantisation error}. This consists of two terms: the loss of accuracy due to values lying outside the absolute range of a format (overflow or underflow) is termed the \textit{clipping error} (or \textit{saturation error}), whereas the loss of accuracy due to values lying between representable numbers is termed the \textit{rounding error}.

We demonstrate the effect quantisation error has for different formats in Figure~\ref{fig:signal_to_noise}. This shows the signal to noise ratio (SNR) of normally distributed values $X \sim \mathcal{N}(0, \sigma^2)$ quantised in FP16 and FP8 as $\sigma$ varies. SNR measures the faithful reproduction of an input (signal) versus the error (noise) introduced, defined as $\mathop{\mathbb{E}}[X^2] / \mathop{\mathbb{E}}[(q(X) - X)^2]$, where $q(\cdot)$ is the quantisation function mapping an input to the nearest representable value.

The heights of the SNR curves reflect the level of rounding error incurred by each format, and the widths reflect the range in which they are free of clipping error. With the exception of subnormal numbers (which slope away on the left-hand-side), the height of each format’s SNR curve is roughly constant. This reflects the fact that exponents are evenly distributed, giving a relative rounding error that is approximately uniform.

\subsection{Trade-offs of low-precision training}
\label{sec:lpt-tradeoffs}

\paragraph{Drawbacks}
The two common 16-bit formats, FP16 and BFLOAT16, offer different trade-offs: FP16 has more precision, but BFLOAT16 has more range. As a result FP16 is more prone to clipping error, requiring careful scaling, and BFLOAT suffers more from rounding error, which in some cases can degrade model accuracy \citep[e.g.][]{Rae2021}.

For FP8 there is a reduction in both range and precision. For range, the same techniques used to train in FP16 are required, and for precision,
the use of FP8 has thus far been restricted to only the inputs of matmul (matrix multiply) operations \citep{Sun19, Noune22, Micikevicius22}, with 3 mantissa bits typically required for weights and activations, and 2 mantissa bits for gradients.

\paragraph{Benefits}
The potential efficiency gains when using low-precision formats are substantial. These include memory usage (often a limiting factor for large models), bandwidth usage (the main overhead for low-arithmetic-intensity ops), compute (the main overhead for high-arithmetic-intensity ops) and cross-device communication (a substantial overhead for distributed training).

\subsection{Low-precision training techniques}
\label{sec:lpt-techniques}

Here we analyse existing techniques for addressing the challenges of low precision training. Table~\ref{tab:method_comparison} provides a summary of their trade-offs and a comparison with unit scaling.

\paragraph{Mixed precision}

Mixed precision is the use of multiple number formats with different bit-widths. This differs from the traditional approach of placing all values in FP32, with \citet{Micikevicius18} showing that most activations, weights and gradients (collectively, \textit{tensors}) can be put in FP16 with no loss in accuracy, with the exception of master weights that are often kept in FP32. Mixed precision training is also possible in BFLOAT16 \citep{Kalamkar19}.

By `training in FP8' we mean that matmuls are performed in FP8 (inputs are cast down to FP8, with outputs in higher precision) with wider formats typically used elsewhere, following the lead of \citet{Sun19, Noune22} and \citet{Micikevicius22}. FP8 reduces both precision and range, and has not generally been used for other operations as matmuls benefit most from using low-precision formats.

Mixed precision training is complementary to unit scaling---all of our experiments use some form of mixed precision.

\paragraph{Loss scaling}

Reduced range in FP16 and FP8 is particularly challenging for the backward pass, where standard model-design practices lead to gradients that risk underflow.

To combat this, \citet{Micikevicius18} have observed that the loss can be multiplied by a scalar to increase the scale of gradients, where weight gradients are then divided by the same scalar in the optimiser. This is valid due to the linearity of the backward pass implicit in the chain rule. Loss scaling is often essential to accurate mixed precision training in FP16 and FP8.

However, there is no theoretical motivation for the choice of loss scale, which instead must be found empirically. This comes with a number of downsides.
Firstly, a hyperparameter sweep must be conducted to find the loss scale value. This can require multiple full runs, as insufficient loss scales may only become apparent later in training. Secondly, it's not clear ahead-of-time what changes require the loss scale to be re-swept. Thirdly, as loss scaling only applies a single, global scaling factor, it has no mechanism to combat differences in scale between gradient tensors. For some models this difference may be too large for effective training.

\paragraph{Automatic loss scaling}

The dynamic adjustment of the loss scale during training is termed \textit{automatic loss scaling} \citep{Kuchaiev2018}. This can remove the need to sweep the initial loss scale, and combats shifts in tensor distributions during training.

The combination of automatic loss scaling and automatic selection of number formats, is termed \textit{automatic mixed precision} \citep{Pytorch23}. Unit scaling doesn't specify tensors' formats, so can be used in systems that automate it.

\paragraph{Per-tensor scaling}

To address the inherent scaling difficulties of FP8 training, \citet{Micikevicius22} propose a per-tensor scaling system, re-scaling locally based on runtime statistics.

Like unit scaling, at the beginning of training this technique may be able to achieve well-scaled tensors throughout the model. However, additional compute, memory, bandwidth and cross-device communication costs may be incurred by the recording of statistics (see Section \ref{sec:discussion} for a more detailed discussion of the potential compute overheads incurred by each of these schemes).

    \section{{Analysis}}
    \label{sec:analysis}
For normally distributed tensors we use the term \textit{scale} to refer to standard deviation. We observe minimal change (relative to the range of our formats) of the mean. Scale therefore characterises the probability of clipping error given a format, as too large or small a scale will lead to values that lie outside of the representable range.

\paragraph{Ideal scaling}
Given we are able to influence the scale of tensors at the start of training, the questions arises---what scale should we aim for? As suggested by Figure~\ref{fig:signal_to_noise}, we argue that unit scale, $\sigma=1$ is a `sweet spot' representing a sensible compromise between several competing factors. We address this question further in Appendix~\ref{app:unit_criterion}.

\paragraph{Is scale predictable?}
The ability to predict the scales of tensors in a deep learning model would give us a powerful tool to address clipping error. This is hard in general, but the problem is simpler at initialisation. Before any training steps, parameters are drawn from known initialisation distributions, so if the input distribution is known, analysis or simulation can derive the scale of each tensor.

A further simplification is to make local distributional assumptions for a single layer in the model and consider the propagation of scale through the model. This permits a methodical analysis: first, characterise the scaling effect of each operation independently; second, propagate scales through the computational graph, forwards and backwards. We provide an example of such analysis in Appendix~\ref{app:scaling_example}.

\paragraph{Scaling at initialisation}
Since the initial distribution of parameters is directly controlled by the model designer, the dominant approach to scaling is to select initial parameter variance to trade off forward and backward pass variance scaling \citep{Glorot10,He15}.

Such schemes were developed to avoid exploding/vanishing gradients in deep multilayer perceptrons. As such, they do not seek to constrain the scale of parameters and parameter gradients. They are also limited to computations where scale factors can be moved into trainable parameters.

\paragraph{Example: BERT \citep{Devlin19}}
BERT's initialisation scheme does not use the rules of \citet{Glorot10}, instead initialising all non-bias parameters from $N(0, (0.02)^2)$. It also adopts a scaling factor from the Transformer \citep{Vaswani17}, which scales the product of activation matrices $Q K^\top$, $Q, K \in \mathbb{R}^{s \times d}$ by $1/\sqrt{d}$.

We instrument the model to record histograms of all tensors at the start and end of training, and plot the results in Figures~\ref{fig:bert_scaling_reg_init} and \ref{fig:bert_scaling_reg_end}. In light of this analysis, we can understand loss scaling as simply enacting a shift of the \textit{gradx} and \textit{gradw} histograms by $\log_2(\textrm{loss scale})$ bits to the right, trading off underflow and overflow globally across gradient tensors.

BERT with loss scaling illustrates the drawbacks of having just three scales: weight initialisation scale, loss scale, and $Q K^\top$ scale. These are not sufficient to centre most tensors' distributions in the representable range.

    \section{{Unit Scaling}}
    \label{sec:unit_scaling}

Based on our analysis of the scaling within typical models and the limitations of existing methods for managing scale, we present \textit{unit scaling}. A model is said to be unit-scaled if its activations, weight and gradients have approximately unit variance at initialisation.

We achieve this by inserting scaling factors into the forward and backward passes. Like loss scaling, our modification of the backward pass still ensures correct gradients up to a constant multiplicative factor. However, unlike loss scaling, unit scaling determines these scales based on a set of rules for each operation, rather than a single hyperparameter to be found empirically, or via an adaptive algorithm.

The scales chosen enable each operation to approximately preserve the variance of its inputs. This effect then propagates through the model, giving global unit-scaling. By concentrating values in approximately the centre of the exponent range at initialisation, we give tensors headroom to potentially shift during training without going out-of-range.

Unit scaling does not address the issue of adapting scales during training. We anticipate that unit scale is sufficient to avoid numerical instability for many models, and observe this in all our experiments. We leave to further work a full investigation of where dynamic re-scaling is required, and how to integrate such a scheme into unit scaling.

\subsection{A framework for scaling computational graphs}

\paragraph{Computational Graphs}

We take our model to be represented by the differentiable function $f_{\textrm{model}}(x_1, \dots, x_m)$, itself a composition of differentiable functions $f_1, \dots, f_n$.

We can describe the structure of such a model using a directed acyclic graph (DAG) denoted $\mathcal{G} = (\mathcal{V}, \mathcal{E})$, with the property that the vertex $v_i \in \mathcal{V}$ corresponds to the function $f_i$ for each $i \in \{1, \dots n\}$, and where the vector-valued output of function $f_a$ used as an input to function $f_b$ is represented by the edge $(v_a, v_b) \in \mathcal{E}$.

This kind of graph is commonly known as a \textit{computational graph}, with vertices as \textit{nodes} and their corresponding functions as \textit{ops}.

\paragraph{Forward and backward graphs}

We refer to the computational graph corresponding to $f_{\textrm{model}}$ as the \textit{forward graph}.

In deep learning we typically apply reverse-mode automatic differentiation to the forward graph to create a second computational graph whose output nodes represent the partial derivatives of the model with respect to its inputs: $\pdv{f\textrm{model}}{x_i}, \; \forall i \in [1..m]$. We call this the \textit{backward graph}.

The backward graph mirrors the structure of the forward graph, but with edge directions reversed. Thus each op $f$ in the forward graph corresponds to a new op $f_{\text{grad}}$ in the backward graph. This op computes the gradient of the model up to $f$ by calculating the product of the incoming gradient $g$ from the previous grad op and the partial derivatives of $f$ evaluated at its inputs: $f_{\text{grad}}(x_1, \dots, x_k, g)_j \triangleq g^{\top} \pdv{f}{x_j} (x_1, \dots, x_k), \; \forall j \in [1..k]$.

\paragraph{Scaled ops}

Given an op $f(x_1, \dots, x_k)$, we define the \textit{scaled op} $f^{*}(x_1, \dots, x_k, \alpha, \beta_1, \dots, \beta_k)$ with \textit{scaling factors} $\alpha, \beta_1, \dots, \beta_k \in \mathbb{R}^+$, such that:
\begin{align*}
    f^* &\triangleq \alpha \cdot f(x_1, \dots, x_k), \\
    f^*_\text{grad}(x_1, .. x_k, g)_i &\triangleq \beta_i \cdot f_{\text{grad}}(x_1, .. x_k, g)_i, \forall i \in [1..k].
\end{align*}

\begin{proposition}
\label{theorem:dynamics}
For any scaled op, there is an equivalent unscaled op with the same training dynamics under a first-order optimiser.
\end{proposition}

We demonstrate this for SGD and Adam in Appendix~\ref{app:theory_dynamics}.

\paragraph{Scaled computational graph}

A scaled computational graph is one where every op $f$ in the forward graph is replaced by a scaled equivalent $f^*$, with the backward graph then generated to produce $f^*_\text{grad}$ for each $f_{\text{grad}}$, using any choice of scaling factors.

If we can show that a scaled computational graph represents a scaled op, by Proposition~\ref{theorem:dynamics}, we are within a reparameterisation of regular training. Unfortunately, this is not true for scaled computational graphs in general, for example $h^*(x) \triangleq x + f^*(x, \alpha, \beta)$ is not a scaled op for some choices of the scaled op $f^*$ and when $\alpha \neq \beta$ (see Appendix~\ref{app:theory_scaled_graph_example}).

\paragraph{Constraint-scaled computational graphs}

We denote the set of edges in the forward graph that are cut-edges\footnote{A cut-edge is an edge in the equivalent undirected graph where the number of connected components increases upon its deletion.} as $\mathcal{C} \subseteq \mathcal{E}$. A constraint-scaled computational graph is a scaled computational graph where we restrict the scaling factors of ops that consume non-cut-edge variables in the following way: for any edge $e \not\in \mathcal{C}$, we require the op consuming the variable $x_e$ to have scaling factors $\alpha = \beta_e$.

\begin{theorem}
\label{theorem:constraintgraph}
A constraint-scaled computational graph itself represents a scaled op.
\end{theorem}
Proven in Appendix~\ref{app:theory_constraint_graph}. This is sufficient to show that we've achieved the property we set out to: valid gradients, up to a constant multiplicative factor.

\subsection{A scaling strategy for unit variance}
\label{sec:scaling_strategy}

\paragraph{Unit scaled computational graphs}

We define a unit-scaled computational graph as an instance of a constraint-scaled computational graph, with scales selected via the following:
\begin{enumerate}[itemsep=-0.2em]
    \item Initially set aside any scale constraints, and calculate the scaling factors that give each op expected unit variance outputs (this process is covered below).
    \item Now resolve any scale constraints by taking each constrained group $\{\alpha, \beta_1, \dots, \beta_l\}$ and selecting the geometric mean $(\alpha \cdot \beta_1 \cdot \ldots \cdot \beta_l)^{\frac{1}{l+1}}$.
\end{enumerate}

This compromise is necessary to ensure valid gradients, but diverges from strict unit scale. In practice though, we observe that the scales going into our geometric mean are often similar enough to preserve approximate unit variance.

\paragraph{Selecting scaling factors}

Assuming unit-scaled inputs to $y = f(x_i, \dots, x_k)$, derive the output scale $\sigma_Y$ and set the forward scaling factor $\alpha = 1/\sigma_Y$. Repeat this process for $x_i^{\prime} = f_{\textrm{grad}}(\dots)_i$, $\forall i \in [1..k]$, to obtain the gradient scale $\sigma_{x_i^\prime}$ and set the backward scaling factor $\beta_i = 1/\sigma_{x_i^\prime}$. (See Table~\ref{tab:ops_compendium} for the scaling factors of common ops.)

Note that our assumption of unit-scaled inputs above is justified by inductive reasoning: we assume that a given op has unit-scaled inputs, which allows us to unit scale its outputs. In this way, unit scale propagates through the graph. The base-cases here are the model's initial inputs, corresponding to parameters and input data. As we initialise parameters to have unit scale, the only extra step we require is to normalise the input data.

\begin{figure*}[tb]
\begin{minipage}[t]{.325\textwidth}
\codefig{projection.py}
\end{minipage}\hfill
\begin{minipage}[t]{.325\textwidth}
\codefig{ffn.py}
\end{minipage}\hfill
\begin{minipage}[t]{.325\textwidth}
\codefig{ffn_scaled.py}
\end{minipage}
\caption{PyTorch examples. \textit{Left:} Scaled projection op, which implicitly constrains $\beta_{X}$. \textit{Center vs Right:} Unscaled vs scaled Transformer FFN layers. Changes: a) initialise weights with unit scale, b) replace unscaled with scaled ops, c) replace residual add with interpolation according to $\tau$, moving the backward pass scale as in Section~\ref{sec:scaling_strategy}. See Figure~\ref{code:appendix} for the implementation of \texttt{scaled} and further ops.}
\label{code:main}
\end{figure*}

\subsection{Weighted addition}
\label{sec:weighted_addition}

For the most part, the scale of tensors at initialisation in unscaled deep learning models does not play a critical role. A notable exception is when tensors of different scales are added, for example residual layers, losses and positional encodings.

If we na\"ively convert these \texttt{add} ops to unit-scaled equivalents, they place equal weight on their inputs, which can be detrimental to performance. We propose using \texttt{weighted\_add} (Table~\ref{tab:ops_compendium}) to resolve this. This introduces new hyperparameters into the model, which can be chosen by design principle, empirically by sweep, or selected to match a reference model (see Appendix~\ref{app:aligning}).

For residual layers, there are existing design principles in literature. We consider the following residual layers based on NF-ResNets \citep{Brock21}:

\textit{default:} $x_{l+1} = x_l + f(x_l)$ (not suitable for unit scaling)

\textit{fixed ($\tau$):} $x_{l+1} = \sqrt{1-\tau} \cdot x_l + \sqrt{\tau} \cdot f(x_l)$

\textit{running-mean:} $x_{l+1} = \sqrt{l/(l\!+\!1)} \cdot x_l + \sqrt{1/(l\!+\!1)} \cdot f(x_l)$

An issue with these weighting rules is that they may produce small gradient scales in the residual branch, which isn't a cut-edge so can't be independently rescaled. To resolve this, we perform a special-case rewrite to replace $\gamma \cdot f(x)$ with  $\mathrm{id}^*(f(\mathrm{id}^*(x, 1, \gamma)), \gamma, 1)$, where $\mathrm{id}^*(x, \alpha, \beta)$ is the scaled identity function. This maintains unit scale for the backward pass $f_{\mathrm{grad}}$, while preserving $\mathcal{G}$ as a scaled op.

\subsection{Recipe}
\label{sec:recipe}

We now outline a high-level recipe for a unit-scaled model:
\begin{enumerate}
    \item Initialise non-bias parameters with unit variance.
    \item Calculate scaling factors for all scaled ops.
    \item Identify non-cut-edges, and constrain the ops consuming them to have $\alpha = \beta$ by taking the geometric mean.
    \item Replace adds with weighted adds.
\end{enumerate}

Unconstrained scaling factors are as outlined in Appendix~\ref{app:ops_compendium}. Identifying cut-edges may sound challenging, but in practice is similar across models. The set of cut-edges commonly contains parameters and any encoder/decoder layers (anything before/after a stack of residual layers). After applying this recipe, training and inference proceed as usual.

To align a unit-scaled model with an existing model, there are some additional considerations. We cover these in Appendix~\ref{app:aligning}. One notable difference is that unit scaled models have different effective optimiser step sizes across their parameters versus unscaled models.\footnote{For instance, a larger effective step size for bias parameters when using unit scaling. \textit{Effective step size} considers the effect of an optimiser update on model output, rather than parameters.} While this difference can be compensated by per-tensor step size modifiers, it means that the training dynamics may be different by default.

\subsection{Example}

\begin{figure}[h!]
\centering
\includegraphics[width=8.2cm]{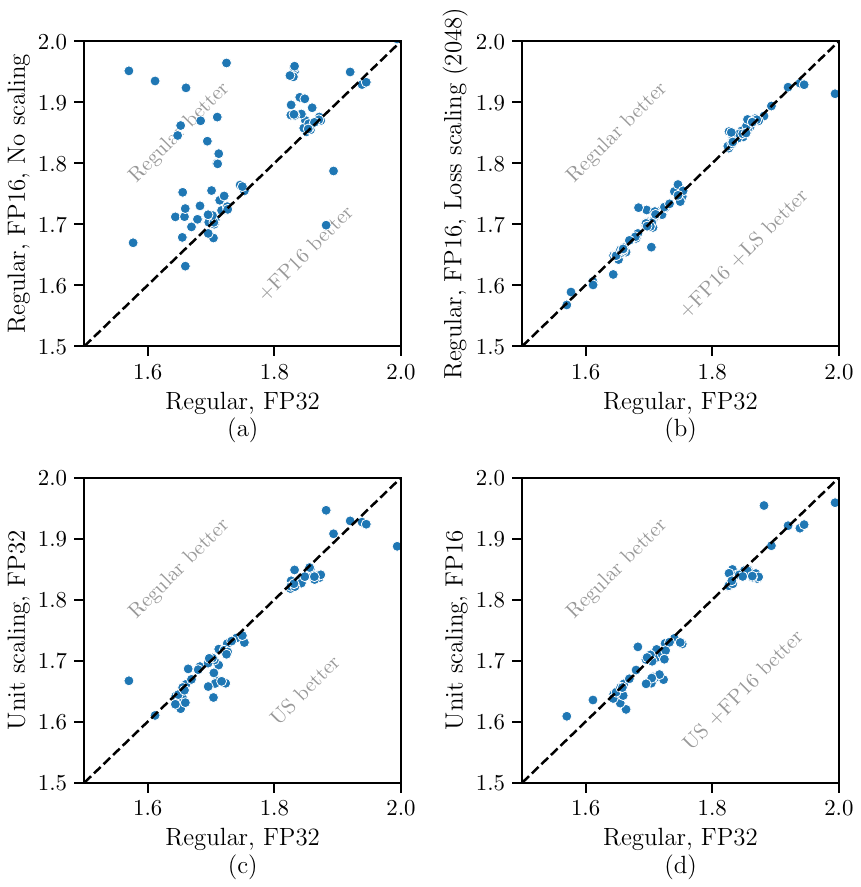}
\caption{Character language modelling, showing validation bits per character over a wide range of models. Each point represents one combination of: \{Conv, RNN, Attention\}, \{Pre, Post, No norm\}, \{Fixed, Running-mean residual\}, \{SGD, Adam\}, \{2, 8 Layers\}. Each point is the best final value over a learning rate sweep.}
\label{fig:char_sweep}
\vspace{-1.2em}
\end{figure}

Using the unit scaling recipe, we first build a scaled op, and then a full scaled layer. Consider a scaled projection op with learnable weights:
\begin{align*}
    \textrm{matmul}^*(X,W) &= \alpha \cdot X \, W \\
    \textrm{matmul}^*_{\textrm{grad}}(X,W,G)_1 &= \beta_1 \cdot G \, W^\top \\
    \textrm{matmul}^*_{\textrm{grad}}(X,W,G)_2 &= \beta_2 \cdot X^\top G \,,
\end{align*}
for input $X \in \mathbb{R}^{b \times m}$, weight $W \in \mathbb{R}^{m \times n}$, output $\mathbb{R}^{b \times n}$ and incoming gradients $G \in \mathbb{R}^{b \times n}$.

Assuming large $b$, $m$, $n$, the analysis of Appendix~\ref{app:scaling_example} gives unconstrained scaling factors $\alpha = m^{-\frac{1}{2}}$, $\beta_1 = n^{-\frac{1}{2}}$, $\beta_2 = b^{-\frac{1}{2}}$. Typically, the edge connecting the weights $W$ is a cut-edge, while the edge connecting in the inputs $X$ is not. Given that assumption, we constrain $\alpha = \beta_1$, satisfied by setting both to the geometric mean of the unconstrained values: $\alpha = \beta_1 = (m \cdot n)^{-\frac{1}{4}}$. We leave $\beta_2$ unchanged.

We show code for the above in Figure~\ref{code:main}, which also gives a scaled layer for the Transformer FFN of Figure~\ref{fig:illustration}.

    \section{{Results}}
    \subsection{Character language modelling}

\paragraph{Experimental Setup} To evaluate unit scaling for multiple model architectures and optimisers, we perform small-scale experiments on WikiText-103 raw character language modelling \citep{Merity17}. We train causal language models, using cross entropy loss during training and evaluate on bits per character (BPC). All models follow the pattern of a Transformer decoder layer \citep{Vaswani17}, with the following variants:

\textit{Sequence layer type}: Attention, RNN and Convolution.

\textit{Norm placement}: PreNorm, PostNorm and NoNorm.

\textit{Residual scaling}: default, fixed and running-mean (as defined in Section~\ref{sec:scaling_strategy}).

Over the product of these settings, we compare the performance of regular (baseline) and unit scaling in both FP32 and FP16. For this, we also evaluate the regular model in FP16 with loss scaling. For full hyperparameters and details, see Appendix~\ref{app:char}.

\paragraph{Results} The above configurations amount to a 2092-run sweep, the results of which are shown in Figure~\ref{fig:char_sweep}. First, these demonstrate the need for scaling when using FP16. This is due to gradient underflow, since loss scaling with a factor of 2048 resolves the issue. Second, they demonstrate that unit scaling, despite changing the training behaviour of the model beyond just numerics, matches or even slightly improves upon baseline performance in almost all cases. Finally, they show that no tuning is necessary when switching unit scaling to FP16.

We also explore the effect of using different residual scaling schemes, with results shown in Figure~\ref{fig:char_residual}. We find that performance is not sensitive to the choice of scheme, and suggest that running-mean or fixed are reasonable choices when using unit scaling.

\subsection{Masked language modelling}
\label{sec:mlm}

\begin{table*}[h!]
\caption{Downstream performance of regular and unit-scaled BERT models. We pretrain 3 models for every \textit{model-method-format} combination, then fine-tune 5 SQuAD v1.1 and 5 v2.0 runs for each (i.e. 15 runs per downstream task). The values shown represent the mean across the 15 runs, with ± indicating the standard deviation across the mean scores of the 3 sub-groups.
† published result from \citet{Devlin19}.
‡ published result from \citet{Noune22}; this model also adds an activation scale alongside the loss scale.}
\label{tab:bert_squad_results}
\centering
\vspace{0.6em}

\begin{tabular}{lllllll}
\toprule
\multirow{2}{*}{Model} &
  \multirow{2}{*}{Method} &
  \multirow{2}{*}{Precision} &
  \multicolumn{2}{c}{SQuAD v1.1} &
  \multicolumn{2}{c}{SQuAD v2.0} \\
 &              &      & \multicolumn{1}{c}{EM}            & \multicolumn{1}{c}{F1}            & \multicolumn{1}{c}{EM}            & \multicolumn{1}{c}{F1}            \\ \hline
\multirow{4}{*}{Base} &
  No Scaling † &
  FP32 &
  80.8 &
  88.5 &
  — &
  — \\ &
  Loss Scaling &
  FP16 &
  80.55 (±0.16) &
  88.19 (±0.16) &
  73.36 (±0.27) &
  76.47 (±0.23) \\
 & Unit Scaling & FP16 & 79.96 (±0.31) & 87.86 (±0.44) & 72.31 (±0.60) & 75.70 (±0.53) \\
 & Unit Scaling & FP8  & 80.15 (±0.18) & 88.04 (±0.12) & 72.28 (±0.02) & 75.67 (±0.01) \\ \hline
\multirow{5}{*}{Large} &
  No Scaling † & FP32 & 84.1 & 90.9 & 78.7 & 81.9 \\
 & Loss Scaling & FP16 & 84.23 (±0.20) & 90.93 (±0.14) & 77.52 (±0.63) & 80.54 (±0.61) \\
 & Loss Scaling ‡ & FP8 & 83.40 (±0.23) & 90.69 (±0.16) & — & — \\
 & Unit Scaling & FP16 & 85.67 (±0.10) & 92.14 (±0.08) & 79.94 (±0.10) & 82.97 (±0.09) \\
 & Unit Scaling & FP8  & 85.22 (±0.03) & 91.77 (±0.10) & 79.29 (±0.31) & 82.29 (±0.29) \\ \bottomrule
\end{tabular}

\end{table*}

\paragraph{Experimental setup}
To evaluate unit scaling against a standard baseline known for challenging numerics, where loss scaling is conventionally required \citep{Lin2020}, we train unit-scaled BERT\textsubscript{BASE} and BERT\textsubscript{LARGE} models.

We use the standard BERT masked language model pretraining objective over English Wikipedia articles, and demonstrate downstream performance on SQuAD v1.1 and SQuAD v2.0 \citep{Rajpurkar2016, Rajpurkar2018}. We follow the unit scaling recipe, along with our guide on aligning a unit scaled model with a regular model (Appendix~\ref{app:aligning}).

Full hyperparameters and details are covered in Appendix~\ref{app:mask}. Note that we do not sweep any additional hyperparameters for our unit-scaled BERT (or character language models) relative to the baselines.

\paragraph{Results}

We report our results in Table~\ref{tab:bert_squad_results}. For unit scaling in FP16, we are able to attain the same performance as the baseline model, and whereas the baseline requires sweeping a loss scale, unit scaling works in all cases out-of-the-box. Due to differences in the effective optimiser step size across parameters (Section~\ref{sec:recipe}), our regular and unit-scaled models aren't exactly equivalent, but deviations in their downstream performance are minor (BERT\textsubscript{BASE} is slightly below the baseline, and BERT\textsubscript{LARGE} is slightly above).

For FP8, we build on the results of \citet{Noune22} who demonstrate the training of loss-scaled BERT in FP8 with no degradation relative to FP16.
We show that the same can also be achieved with unit scaling, with no additional techniques required to make FP8 work over FP16---we simply quantise our matmul inputs into FP8 and are able to train accurately. These results represent the first time BERT\textsubscript{BASE} or BERT\textsubscript{LARGE} have been trained in either FP16 or FP8 without requiring a form of loss scaling.

To highlight the precise effects of unit scaling, we show histograms for activations, weights and gradients for unit-scaled FP16 BERT. These can be found in Figures~\ref{fig:bert_scaling_us_init},~\ref{fig:bert_scaling_us_end}, alongside equivalent plots for a regular FP16 BERT.

The code used in these experiments can be found at \url{https://github.com/graphcore-research/unit-scaling-demo}, alongside a separate notebook implementing a unit-scaled NanoGPT model. We recommend this resource for those looking to understand unit scaling through a simple example implementation.

For those interested in using unit scaling in their own models, we also provide a PyTorch library: \url{https://graphcore-research.github.io/unit-scaling}. The documentation includes a practical guide to developing and optimising a unit-scaled model. This implementation should be considered a definitive reference for unit scaling.

    \section{{Related Work}}
    \paragraph{Variance scaling analysis}
\citet{Klambauer17} and \citet{Peiwen22} propose activation functions that encourage unit-variance activations and gradients, which are complementary to unit scaling. \citet{He16} introduce residual networks, using skip connections and explicit normalisation to stabilise forward and backward passes. Variants on normalisation~\citep{Ioffe15,Ba16,Labatie21,Salimans16}  are complementary to unit scaling, which considers the norm of the gradients as well as activations and does not constrain activation norms after initialisation. Alternative residual schemes~\citep{Zhang19,Brock21} can be incorporated into unit-scaled models, although the residual layer output variance should not be allowed to grow with depth.

The reparameterisation implied by unit scaling is also used by \citet{Jacot18}, later broadened by \citet{Yang20} and exploited by \citet{Yang22} in their work analysing the training behaviour of deep networks. Motivated by low-precision computation rather than training dynamics, unit scaling applies scaling factors locally throughout the compute graph, but the effect on training hyperparameter scaling is similar.

\paragraph{FP8 inference}
Although there has been little hardware support for FP8 training, accelerated 8-bit inference is increasingly common via the use of integer quantisation \citep{Jacob18} to the INT8 format. This process typically results in degraded accuracy, requiring additional techniques such as quantisation-aware training (see \citet{Nagel21} for a thorough discussion on this topic). Though recent efforts have been made to improve efficient INT8 quantisation \citep{Yao22, Park22, Dettmers22, Xiao22}, the use of FP8 enables accelerated inference in the same format as training, promising a substantial improvement in the simplicity and accuracy of 8-bit inference \citep{Kuzmin22}.

    \section{{Discussion}}
    \label{sec:discussion}
\paragraph{Compute overhead} Unit scaling relies solely on the addition of scaling operations of the form $\gamma \cdot X$, where $\gamma$ is a fixed scalar and $X$ is a tensor. These scaling factors can be fused into the preceding ops (e.g. via \texttt{torch.jit}, \texttt{torch.compile}  or \texttt{jax.jit}). By doing this we observe that the increase in memory-access cost is negligible. For models with reasonably large hidden sizes, the compute overhead is also minimal. For example, the FLOPs required to train our unit-scaled BERT\textsubscript{LARGE} are only 0.2\% greater than the baseline (explained further in Appendix~\ref{app:overhead}). Basic loss scaling operates on a similar principle, and only introduces a single scaling factor. From this we conclude that both techniques have low overall overhead, assuming a fused implementation.

Automatic loss scaling has an additional feature which increases overhead: its requirement to occasionally discard batches. This assumes that re-scaling is determined by tracking gradient overflows (the standard approach, as used in \citet{Pytorch23}). When overflows occur, batches must not be used to update parameters. The overhead of dropping batches is tolerable for FP16 but may not be for FP8 \citep{Micikevicius22}.

Proposed automatic per-tensor scaling schemes take a different approach, and have potential to add overhead in other areas (how much depends largely on software and hardware characteristics). \citet{Micikevicius22} reject scaling based on gradient overflows, instead opting for heuristics based on properties of the tensors being scaled. Their preferred training heuristic is not specified, but for inference they choose between max, percentile, and minimum MSE methods. These approaches trade-off overhead for accuracy. At one extreme, max is likely easy to fuse but may be distorted by outliers; at the other extreme minimum MSE may be more robust but is challenging to implement efficiently (e.g. \citet{Sakr2022}). Distributed training adds further challenges, potentially requiring the communication of statistics across devices to keep scales synchronised.

It remains to be seen whether effective automatic scaling methods can be implemented efficiently given these complexities. This will likely be an important future research objective. In contrast unit scaling, with fixed precomputed scaling factors, offers a simpler alternative.

\paragraph{Broader impact}
The potential for unit scaling to simplify the use of 8-bit number formats may lead to increased adoption, and in turn facilitate training larger models. At scale, new capabilities emerge \citep{Wei22}, potentially exacerbating known harms \citep{Weidinger21} such as toxicity \citep{Nadeem20}, misinformation \citep{Lin2021}, privacy concerns \citep{Carlini21} and environmental damage \citep{Strubell19}. To mitigate these outcomes, a variety of methods have been proposed, including reinforcement learning from human \citep{Ouyang22} or AI \citep{Bai22} feedback, anti-experts \citep{Liu2021} and baked-in safety models \citep{Xu20}, all of which are applicable to unit-scaled models.

\paragraph{Conclusion}
We have demonstrated that unit scaling addresses the complexities of low-precision training, providing a simpler and more granular solution. This is demonstrated by our training of BERT\textsubscript{LARGE} for the first time without loss scaling, in FP16 and even FP8. The community's transition to FP8 training will see new capabilities emerge as a result of improved efficiency, and this transition can be accelerated by unit scaling.

\section*{Acknowledgements}
We would like to thank the following people for their contributions to the paper at the various stages of its development: Daniel Justus, Alberto Cattaneo, Andrew Fitzgibbon, Paul Balanca, Luke Prince, Ivan Chelombiev, Luka Ribar and Zach Eaton-Rosen.

\bibliographystyle{plainnat}
\bibliography{references}

\clearpage
\appendix
\renewcommand\thefigure{A.\arabic{figure}}
\renewcommand\theHfigure{A.\arabic{figure}}
\setcounter{figure}{0}
\renewcommand\thetable{A.\arabic{table}}
\renewcommand\theHtable{A.\arabic{table}}
\setcounter{table}{0}
\section{Floating point format specification}
\label{app:float_spec}

\begin{table}[h!]
    \caption{Common floating point formats for deep learning. $E$ refers to the number of exponent bits, and $M$ the number of mantissa bits of a given format. \textit{Max exp.} and \textit{Min exp.} refer to the maximum and minimum values that can be represented by the exponent, excluding special values. E5 (a) and E4 (a) refer to the FP8 formats introduced by \citet{Noune22}, whereas E5 (b) and E4 (b) refer to those introduced by \citet{Micikevicius22}}
    \label{tab:fp_formats}
    \vspace{1em}
    \centering
\begin{tabular}{lllll}
\toprule
Format          & $E$ & $M$ & Max exp. & Min exp. \\
\midrule
FP32            & 8   & 23  & 127      & -126     \\
TF32            & 8   & 10  & 127      & -126     \\
BFLOAT16        & 8   & 7   & 127      & -126     \\
FP16            & 5   & 10  & 15       & -14      \\
FP8 E5 (a)      & 5   & 2   & 15       & -15      \\
FP8 E5 (b)     & 5   & 2   & 15       & -14      \\
FP8 E4 (a)      & 4   & 3   & 7        & -7       \\
FP8 E4 (b)       & 4   & 3   & 8        & -6       \\
\bottomrule
\end{tabular}
\end{table}

\section{Proposed FP8 formats}
\label{app:fp8_formats}

Here we analyse the recently-proposed FP8 formats. We cover two proposals for 8-bit floating point formats \citep{Noune22, Micikevicius22} (other proposals include \citet{Tesla21, Kuzmin22}), each of which introduce one format with 4 exponent bits and a second format with 5. We refer to these here as E4 and E5 respectively (with the implication that the remaining bits represent the sign and mantissa).

To compensate for the low number of representable values, all of the proposed formats except the \citet{Micikevicius22} E5 format deviate from the IEEE 754 standard by reducing the number of special values available. Both \citet{Noune22} formats also increment the IEEE 754 bias by one. This slightly alters the maximum and minimum (absolute normal) values that each can represent.

FP8 formats in the literature are sometimes presented as having an explicit \textit{bias} value, to be defined by the user \citep{Noune22, Kuzmin22}. The bias is subtracted from the exponent, just as in the IEEE 754 standard. This approach is equivalent to multiplying by $2^{-\textrm{bias}}$, and hence is no different from using a scaling factor to control the range of values represented. \citet{Micikevicius22} explore both interpretations, with a preference for the scaling-factor viewpoint which aligns better with software implementations, whereas the exponent-bias viewpoint is more hardware aligned and in practice is likely to restrict bias values to integers.

These caveats aside, the proposed FP8 formats do not differ significantly from a standard-compliant 8-bit format.

\section{Is unit standard deviation the correct criterion?}
\label{app:unit_criterion}

Here we justify the position that aiming for unit standard deviation of normally distributed tensors at initialisation is a sensible strategy.

When considering the scale of a floating-point tensor, we aim to keep absolute values within the upper and lower absolute normal bounds defined by a given format.

To analyse the \textit{absolute} values generated by a normal distribution, we instead consider a folded normal distribution with zero mean and unit variance. Here, the central 90\% of all probability mass falls within $[2^{-4}, 2^1]$.

As a point of comparison, for an IEEE 754 float the absolute range of normal values we can represent is approximately $\left[2^{2^{E-1}}, 2^{2-2^{E-1}}\right]$, giving a centre-point (in log-space) of $2^1$. From the perspective of clipping error, one might suggest scaling values to be as close as possible to this point, as we are equidistant from the upper and lower boundaries.

Hence, we can conclude that unit standard deviation will concentrate most values very near to, but slightly below the centre of the numerical range.
Whether centrality within the normal floating-point range is the correct criterion for normally-distributed tensors during the training of deep learning models is a much harder question to answer.

In favour of sub-central scaling, is the argument that the subnormal values provides us with extra range at the lower end of the spectrum, albeit with reduced precision. Additionally, underflow in deep learning models tends to be less detrimental to training than overflow.

In favour of super-central scaling, is the argument that we might expect values such as gradients to decrease in magnitude during the course of training (our results in Section \ref{app:bert_scaling} suggest that this is true for BERT's \textit{gradw} values, though not for \textit{gradx}), and so we ought to up-scale values to compensate.

In light of these arguments, we argue that in situations where we can control scale, aiming for unit scaling is a sensible compromise. If we wished to precisely align the 90\%-probability-mass range with the centre point calculated above, we might aim for a slightly larger scale. But given the confounding factors outlined, the difference is small enough that $\sigma = 2^0$ is still a strong choice, and keeps us aligned with other techniques in the literature with the same aim (e.g. \citet{Glorot10}).

\section{Unit scaling and emergent outliers}
\label{app:emergent_outliers}
Recent work on inference quantisation for large language models ($>$1B parameters) has highlighted the importance of special techniques for accommodating \textit{outliers}. These are large-magnitude values concentrated in particular sequence-elements \citep{Bondarenko2021} or feature-dimensions \citep{Dettmers22}, emerging as model size increases.

The main difficulty with accommodating tensors with outliers is that ``a single outlier can reduce the quantisation precision of all other values'' \citep{Dettmers22}. These outliers have been shown to degrade INT8 quantisation accuracy at the 6.7B parameter model size and above, which leads to a key question: what impact do we expect outliers to have for unit scaling when applied to models of this size?

Firstly, we do not expect unit scaling to have a significant effect on the \textit{magnitude} of outliers. This is because outliers occur in activation tensors, and these typically have a similar scale in unit and non-unit-scaled models (primarily due to the frequent presence of layernorms, which control scale).

However, we still expect unit scaling to be less impaired by outliers than the examples seen in recent literature. The key consideration here is that unit scaling is a \textit{training} method and uses \textit{floating-point} formats. In contrast, the literature on emergent outliers has all been in the integer quantisation setting.

Integer formats lack the dynamic range for training \citep{Noune22}, and the same problem arises in the presence of outliers. We anticipate that using FP8 over INT8 will mitigate the difficulties presented to unit scaling by outliers. An analysis of the relative SNRs of the formats is insightful:

We first make some assumptions about the problem setting. We take the work of \citet{Dettmers22} as our starting point, who show that the median outlier magnitude is 60 as accuracy begins to degrade. The distribution of non-outlier values is not clear, though the authors define non-outliers to have a magnitude of $<6$. Hence, we assume that these have remained approximately unit-scaled.

To represent values in INT8 we will assume that they are scaled throughout such that outliers are (just) within the range of the format. This involves dividing by the expected maximum outlier value, and multiplying by the maximum INT8 value (127). We will assume a maximum outlier value of $3 \times$ the median, giving a scaling of $127/(3 \times 60)$. To represent values in FP8 (E4) we do not need to re-scale values to accommodate outliers as the maximum FP8 E4 value is already larger than the maximum outlier, at 240.

Having scaled INT8 to accommodate outliers, the key question is what effect this has on the representation of non-outlier values. As observed in the literature, the “range of the quantisation distribution is too large so that most quantisation bins are empty and small quantisation values are quantised to zero, essentially extinguishing information” \citep{Dettmers22}.

We model this scenario, calculating an SNR for the non-outlier values of only 2.03 (this raises to 14.8 if we scale for the median outlier rather than the max). In contrast, the SNR calculated in FP8 E4 is 635x higher at $1.29 \times 10^3$. This is due to the exponential distribution of values in floating-point formats, which gives it a small number of large values (suitable for outliers) and a large number of small values (suitable for non-outliers).

\begin{figure}[tbp]
    \centering
    \includegraphics[width=8.2cm]{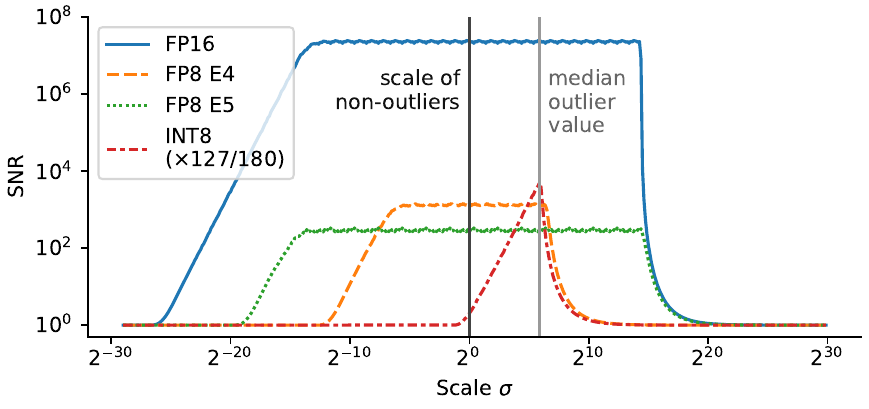}
    \caption{The signal to noise ratio (SNR) of a quantised normal distribution, as a function of the distribution’s scale. This plot is the same as Figure~\ref{fig:signal_to_noise}, but with the addition of scaled INT8 quantisation and vertical lines for outliers and non-outliers.}
    \label{fig:signal_to_noise_outliers}
\end{figure}

This can be observed in Figure~\ref{fig:signal_to_noise_outliers}, where we plot the SNR for this INT8 quantisation applied to a normally distributed tensor across different scales. Although INT8 gives a good representation of the outlier values (as does FP8 E4), the non-outlier values have low signal. One challenge for FP8 is the scenario in which outlier magnitude increases; in this case we would have to either re-scale or switch to the less precise E5 format.

Another way of viewing this is to look at the number of quantisation bins each format makes use of in this setting. For INT8 the lower 95\% of non-outlier values are assigned to just 3 out of 256 quantisation bins. In contrast, for FP8 90 bins are utilised.

This modelling gives us cause for optimism when applying unit scaling in the presence of outliers, though we acknowledge there may still be challenges.

\section{Theoretical results}
\label{app:theory}

\subsection{Example -- scaling analysis}
\label{app:scaling_example}

We reproduce a simple version of the scaling analysis of \citet{Glorot10}, for a multilayer perceptron (MLP).

Consider an MLP which transforms inputs $X_0$ to outputs $X_L$ using $X_{l+1} = f(X_l W_l)$ for $l \in [0, \dots, L-1]$, where $f(\cdot)$ is an elementwise activation function. We separate the analysis of a single layer into $Z = X W$ and $Y = f(Z)$.

\paragraph{Projection}
First, $Z = X W$, where $Z \in \mathbb{R}^{b \times n}$, $X \in \mathbb{R}^{b \times m}$, $W \in \mathbb{R}^{m \times n}$, and $X$, $W$ each have independently distributed elements with zero mean and variance $\sigma_X^2$ and $\sigma_W^2$ respectively. The values in $Z$ follow $Z_{ik} = \sum_j X_{ij} W_{jk}$, which is a sum over $m$ uncorrelated products, each with variance $\sigma_X^2 \sigma_W^2$. Then, by the variance of an independent sum, the output variance $\sigma_Z^2 = m\, \sigma_X^2 \sigma_W^2$.

When computing the partial derivative of a scalar loss $L$ with respect to $X$, $\nabla_X L = (\nabla_Z L) \, W^{\top}$, assuming $\nabla_Z L$ is zero mean with variance $\sigma_{\nabla_Z L}^2$ and is not correlated with $W$,\footnote{This is likely to be a very bad assumption, since $W$ was used to generate $Z$ and therefore $\nabla_Z L$. But it is hard to avoid this assumption without doing a global analysis of the model.} then by same reasoning as above $\sigma_{\nabla_X L}^2 = n\, \sigma_{\nabla_Z L}^2 \,\sigma_W^2$. And again $\sigma_{\nabla_W L}^2 = b\, \sigma_{\nabla_Z L}^2 \, \sigma_X^2$.

\paragraph{Activation}
Consider $f(Z) = \mathrm{relu}(Z) = \mathrm{max}(Z, 0)$, with $Z \sim \mathcal{N}(0, 1)$. Then, in the forward pass $\mathrm{P}(f(Z)=y) = \frac{1}{2} \delta(y) + H(y) \cdot \mathrm{P}_{\mathcal{N}}(y)$, where $\mathrm{P}_{\mathcal{N}}(\cdot)$ is the pdf of a standard normal distribution, and $H(\cdot)$ is the Heaviside step function. This gives variance $\sigma_Y^2 = \frac{1}{2} (1 - 1/\pi)$. In the backward pass, $\mathrm{P}(\nabla_Z L = z^\prime) = \frac{1}{2} \delta(z^\prime) + \frac{1}{2} \mathrm{P}_{\mathcal{N}}(z^\prime)$, with variance $\sigma_{\nabla_Z L}^2 = \frac{1}{2}$.

\citet{He15} note that the activation function can break the local distributional assumption for the first step: for example, the ReLU function $f(Z) = \mathrm{max}(Z, 0)$ does not produce zero mean output, invalidating our previous assumption on $X_l$. However, the corrections for such invalid assumptions are often small, and can be ignored for sake of expedience, permitting local scaling analysis.

For an example of extending scale analysis to training, \citet{Huang20} consider the training dynamics of a Transformer under Adam, using this to derive an initialisation scheme that avoids vanishing updates.

\subsection{Proofs in support of Proposition~\ref{theorem:dynamics}}
\label{app:theory_dynamics}

For two common choices of optimiser, SGD and Adam, we show that there is an unscaled model with identical training dynamics as any unit-scaled model.

\subsubsection{SGD}

We define a model as an op with scalar output and a subset of inputs denoted as trainable parameters $\theta_i$, written $f(\theta_{i \in 1 \dots n}, x_{j \in 1 \dots k})$.

A \textit{training trajectory} is defined as a sequence $\theta^{(t)}_i$ for all parameters in a model, given initial settings $\theta^{(0)}_i$ and optimiser. For SGD,
\begin{align*}
    \theta^{(t+1)}_i
    &= \theta^{(t)}_i - \eta \frac{\partial f(\dots)}{\partial \theta_i} \,,\\
    &= \theta^{(t)}_i - \eta f_{\mathrm{grad}}(\dots, 1)_i \,,
\end{align*}
where $\eta$ is a constant learning rate hyperparameter. We define the trajectory under a scaled op similarly, using $f^*_{\mathrm{grad}}$:
\begin{equation*}
    \theta^{*,(t+1)}_i = \theta^{*,(t)}_i - \eta f^*_{\mathrm{grad}}(\dots, 1)_i \,.
\end{equation*}

\begin{proposition}
    For any scaled op with training trajectory $\theta^{*,(t)}_i$ under SGD, there exists an equivalent unscaled op with training trajectory $\theta^{(t)}_i = \sqrt{\alpha / \beta_i} \cdot \theta^{*,(t)}_i$.
\end{proposition}

We consider the evolution of the following unscaled op under SGD on $\theta$:
\begin{align*}
    \hat{f}(\theta_{i \in 1\dots n}, x_{j \in 1\dots k})
    &\triangleq \alpha \cdot f(\sqrt{\beta_i / \alpha} \cdot \theta_{i \in 1\dots n}, x_{j \in 1\dots k}) .
\end{align*}
Applying the chain rule to obtain gradients,
\begin{equation*}
    \frac{\partial \hat{f}(\theta_{i^\prime \in 1\dots n}, \dots)}{\partial \theta_i}
    = \alpha \cdot \sqrt{\beta_i / \alpha} \cdot \frac{\partial f(\sqrt{\beta_{i^\prime} / \alpha} \cdot \theta_{i^\prime \in 1\dots n}, \dots)}{\partial \theta_i} .
\end{equation*}
Substituting to get the evolution of $\theta_i$ under SGD,
\begin{equation*}
    \theta^{(t+1)}_i = \theta^{(t)}_i - \eta \cdot
    \sqrt{\alpha / \beta_i} \cdot \beta_i \cdot \frac{\partial f(\sqrt{\beta_{i^\prime} / \alpha} \cdot \theta^{(t)}_{i^\prime \in 1\dots n}, \dots)}{\partial \theta_i} .
\end{equation*}
We can now use the define $\theta^*$ as follows and obtain
\begin{align*}
    \theta^*_i &\triangleq \sqrt{\beta_i / \alpha} \cdot \theta_i \,,\\
    \theta^{*,(t+1)}_i &= \theta^{*,(t)}_i - \eta f^*_{\mathrm{grad}}(\theta^{*,(t)}_{i \in 1\dots n}, \dots, 1)_i .
\end{align*}
Therefore if the initial condition $\theta^{(0)}_i = \sqrt{\alpha / \beta_i} \cdot \theta^{*,(0)}_i$ is satisfied, then $\theta^{(t)}_i = \sqrt{\alpha / \beta_i} \cdot \theta^{*,(t)}_i$ thereafter.

\subsubsection{Adam}

As noted by \citet{Kingma15}, Adam is invariant to diagonal rescaling of the gradients. Defining the function $\mathrm{adam}$ that computes a single update thus:
\begin{equation*}
    \theta^{(t+1)} = \mathrm{adam}(\theta^{(t)}, \frac{\partial f}{\partial \theta}) \,,
\end{equation*}
invariance to diagonal rescaling gives
\begin{equation*}
    \mathrm{adam}(\theta^{(t)}, \frac{\partial f}{\partial \theta}) = \mathrm{adam}(\theta^{(t)}, s \odot \frac{\partial f}{\partial \theta}) \,,
\end{equation*}
for any positive-valued scaling vector $s \in \left(\mathbb{R}^+\right)^{|\theta|}$ that is constant over all timesteps $t$.

\begin{proposition}
    For any scaled op with training trajectory $\theta^{(t)}_i$ under Adam with $\epsilon=0$, there exists an equivalent unscaled op with the same training trajectory.
\end{proposition}

Consider the unscaled op $\hat{f}(\dots) = \alpha \cdot f(\dots)$. This follows the trajectory
\begin{equation*}
    \theta^{(t+1)}_i = \mathrm{adam}(\theta^{(t)}_i, \alpha \cdot \frac{\partial f}{\partial \theta_i}) .
\end{equation*}

Now consider the scaled op $f^*$ with the same $\alpha, f$. This follows:
\begin{align*}
    \theta^{*,(t+1)}_i &= \mathrm{adam}(\theta^{*,(t)}_i, \beta_i \cdot \frac{\partial f}{\partial \theta_i}) ,\\
    &= \mathrm{adam}(\theta^{*,(t)}_i, \left(\frac{\beta_i}{\alpha}\right) \cdot \alpha \cdot \frac{\partial f}{\partial \theta_i}) .
\end{align*}
Therefore if $\theta^{*,(0)} = \theta^{(0)}$, we conclude $\theta^{*,(t)} = \theta^{(t)}$.

\subsection{Example -- a scaled computational graph does not necessarily represent a scaled op}
\label{app:theory_scaled_graph_example}

  Let $f(x_1, \dots, x_n)$ be an unscaled operation with values in $\mathbb{R}^n$ and consider the scaled computational graph defined by $x + f^*(x, \alpha, \beta_1, \dots, \beta_n)$. If this scaled computational graph represented a scaled op $h^*(x_1, \dots, x_n)$ for some function $h(x_1, \dots, x_n)$, there would exist scalars $\alpha', \beta'_1, \dots, \beta'_n$ such that:
\begin{align*}
    \alpha' h(x) &= x + f^*(x, \alpha, \beta) \,, \\
    \beta'_i g^\top \frac{\partial h(\dots)}{\partial x_i} &= g_i + f^*_{\mathrm{grad}}(x, \alpha, \beta, g)_i \; \forall i \in \left\{ 1, \dots, n \right\}.
\end{align*}

Consider $f(x) = x^2$, so that
\begin{align*}
    f^*(x, \alpha, \beta) &= \alpha \cdot x^2 \,, \\
    f^*_{\mathrm{grad}}(x, \alpha, \beta, g)_i &= 2\beta_i \cdot x_i \cdot g_i \; \forall i \in \left\{ 1, \dots, n \right\}.
\end{align*}

This implies $$\frac{\beta'_i}{\alpha'} \cdot g_i \cdot \left(1 + 2\alpha x_i \right) = g_i + 2\beta_i \cdot g_i \cdot x_i \; \forall i \in \left\{ 1, \dots, n \right\}.$$
Assuming $g_i \neq 0$, in the case $\alpha \neq \beta_i$ these two expressions cannot be made to match by any choice of $(\alpha', \beta'_i)$. Therefore the scaled graph does not implement a scaled op.

\subsection{Proof of Theorem~\ref{theorem:constraintgraph}}
\label{app:theory_constraint_graph}

We first define how a computational graph represents an op. Then we show that an unscaled graph correctly represents an unscaled op. Finally, we proceed to show that a constraint-scaled graph with a single output correctly represents a scaled op.

\paragraph{Graph -- op}
We adopt a generalisation of the earlier definition of an op, to permit multiple outputs. An op defines mappings from $k$ vector-valued inputs to $m$ vector-valued outputs via $f(x_{i \in 1\dots k})_{j \in 1\dots m}$, and corresponding gradient mappings,
\begin{equation*}
    f_{\mathrm{grad}}(x_{i' \in 1\dots k}, g_{j' \in 1\dots m})_i
    \triangleq
    \sum_j g_j^\top \frac{\partial f(x_{i' \in 1\dots k})_j}{\partial x_i} \,.
\end{equation*}
We use $f_{\mathcal{G}}$ to denote the \textit{graph op} represented by the computational graph $\mathcal{G}$. To evaluate the function and the vector Jacobian product $f_{\mathrm{grad},\mathcal{G}}$, we assign inputs and outputs to edges in the graph.\footnote{It is often natural to assign inputs and outputs to nodes, but we use edges in our analysis for notational convenience. Such edges imply the existence of `dummy' nodes.} Define a list of input edges, $\mathrm{in}_{i \in 1\dots k} \in \mathcal{E}$, and output edges, $\mathrm{out}_{j \in 1\dots m} \in \mathcal{E}$.

Define the \textit{forward value} of an edge using $z : \mathcal{E} \to \mathbb{R}^{(\cdot)}$, via the recursive relations:
\begin{align*}
    z(\mathrm{in}_i) &\triangleq x_i \,,\\
    z((u,v)) &\triangleq
        f_u(\left\{z((w,u)) \mid (w,u) \in \mathcal{E} \right\})_v, \\
    f_{\mathcal{G}}(x_{i \in 1\dots k})_j &\triangleq z(\mathrm{out}_j),
\end{align*}
where $f_u(\dots)_v$ evaluates node $u$'s output corresponding to the edge $(u, v)$.

Similarly, define the \textit{backward value} of an edge using $h : \mathcal{E} \to \mathbb{R}^{(\cdot)}$ via:
\begin{align*}
    h(\mathrm{out}_j) &\triangleq g_j,\\
    h((u,v)) &\triangleq f_{\mathrm{grad},v}(
        \left\{z((u',v))\right\},
        \left\{h((v,r))\right\})_u, \\
    f_{\mathrm{grad},\mathcal{G}}(\dots, g_{j \in 1\dots m})_i &\triangleq h(\mathrm{in}_i),
\end{align*}
where $f_{\mathrm{grad},v}(\dots)_u$ evaluates the grad op for node $v$ for the input $x_{v,u}$ corresponding to the edge $(u, v)$. Note that we use the shorthand $\left\{z((u',v))\right\}$ to denote $\left\{z((u',v)) \mid (u',v) \in \mathcal{E}\right\}$.

\paragraph{Unscaled graph -- op}
To show that $(f_{\mathcal{G}}, f_{\mathrm{grad},\mathcal{G}})$ represent an op, we must show they are consistent with the definition of $f_{\mathrm{grad}}$. We expand the backward value using the definition of $f_{\mathrm{grad},v}$,
\begin{equation*}
    h((u,v)) = \sum_w h((v,w))^\top
        \frac{\partial f_v(\left\{z((u',v))\right\})_w}{\partial x_{v,u}}.
\end{equation*}
Using the base case for $h(\mathrm{out}_j)$ and the chain rule,
\begin{align*}
    h((u,v)) &= \sum_w \left(\sum_q h((w,q))^\top \frac{\partial f_w(\dots)_q}{\partial x_{w,v}}\right)^\top
    \frac{\partial f_v(\dots)_w}{\partial x_{v,u}} ,\\
    h((u,v)) &= \sum_j g_j^\top
        \frac{\partial f_{\mathcal{G}, v}(\dots)_j}{\partial x_{v,u}}.
\end{align*}
Therefore $h(\mathrm{in}_i)$ gives the correct gradient, so $\mathcal{G}$ correctly represents an op.

\paragraph{Constraint-scaled graph -- scaled op}
Again, generalising the earlier definition to multiple outputs,
\begin{align*}
    f^*(x_{i \in 1\dots k})_j
    &\triangleq
        \alpha \cdot f(x_{i \in 1\dots k})_j \,,\\
    f^*_{\mathrm{grad}}(x_{i' \in 1\dots k}, g_{j' \in 1\dots m})_i
    &\triangleq
        \beta_i \cdot\sum_j g_j^\top \frac{\partial f(x_{i' \in 1\dots k})_j}{\partial x_i} \,.
\end{align*}
Note that all outputs are scaled using a single value $\alpha$. Using the same definitions for $z$ and $h$,
\begin{align*}
    h((u,v)) &= \beta_{v,u} \sum_w h((v,w))^\top
        \frac{\partial f_v(\left\{z((u',v))\right\})_w}{\partial x_{v,u}} ,\\
    &= \frac{\beta_{v,u}}{\alpha_v} \sum_w h((v,w))^\top
        \frac{\partial f^*_v(\left\{z((u',v))\right\})_w}{\partial x_{v,u}}.
\end{align*}

In order to apply the chain rule here, we must first deal with the scale ratio $\frac{\beta_{v,u}}{\alpha_v}$. To do this, we define the \textit{unscaled backward value}, $\hat{h}$, in terms of a single reachable output $\mathrm{out}$ and a rescaling function $s : \mathcal{E} \times \mathcal{E} \to \mathbb{R}$, thus:
\begin{align*}
    \hat{h}((u,v)) &\triangleq \frac{h((u,v))}{s((u,v), \mathrm{out})} ,\\
    s(a, b) &\triangleq \prod_{(u,v) \in \mathcal{E}^{\mathrm{cut}(a, b)}} \frac{\beta_{v,u}}{\alpha_v} ,
\end{align*}
where $\mathcal{E}^{\mathrm{cut}(a, b)}$ is the set of edges where, after the removal of any one, there is no path connecting the head of $a$ and the head of $b$ in $\mathcal{G}$. We observe this property for adjacent edges:
\begin{equation*}
    \frac{s((v,w), \mathrm{out})}{s((u,v), \mathrm{out})}
    = \begin{cases}
        \frac{\alpha_v}{\beta_{v,u}} & \textrm{ if $(u,v)$ is a cut-edge} \\
        1 & \textrm{ otherwise}
    \end{cases},
\end{equation*}
which follows directly from the definition of $s$. Now we substitute into our grad,
\begin{align*}
    \hat{h}((u,v)) &= \sum_w
        \gamma(u, v, w) \cdot
        \hat{h}((v,w))^\top
        \frac{\partial f^*_v(\dots)_w}{\partial x_{v,u}} ,\\
    \gamma(u, v, w) &\triangleq
        \frac{\beta_{v,u}}{\alpha_v}
        \cdot
        \frac{s((v,w), \mathrm{out})}{s((u,v), \mathrm{out})}.
\end{align*}
Consider two cases:

\textit{Case 1: $(u,v)$ is not a cut-edge.} The rules of constraint-scaled computation graphs ensure $\beta_{v,u} = \alpha_v$. From the aforementioned property, $s((u,v), \mathrm{out}) = s((v,w), \mathrm{out})$. So we conclude $\gamma(u, v, w) = 1$.

\textit{Case 2: $(u,v)$ is a cut-edge.} From the same property, we conclude $\gamma(u, v, w) = 1$.

Since in either case, $\gamma(u, v, w) = 1$, we can simplify:
\begin{equation*}
    \hat{h}((u,v)) = \sum_w
        \hat{h}((v,w))^\top
        \frac{\partial f^*_v(\dots)_w}{\partial x_{v,u}} ,\\
\end{equation*}
which is the correct form for the chain rule and induction from the base case as previously, noting that $s(\mathrm{out}, \mathrm{out}) = 1$ so $\hat{h}(\mathrm{out}) = g$. We can therefore conclude that $\hat{h}$ gives true gradients and:
\begin{align*}
    \hat{h}((u,v)) &= g^\top
        \frac{\partial f_{\mathcal{G}, v}(\dots)}
        {\partial x_{v,u}} ,\\
    h((u,v)) &= s((u,v), \mathrm{out}) \cdot \hat{h}((u,v)).
\end{align*}
So $\mathcal{G}$ represents a scaled op with $\beta_i = s(\mathrm{in}_i, \mathrm{out})$.

\section{Constraint-scaled computational graphs for other schemes}

For sake of comparison, it can be instructive to consider other scaling schemes within the constraint-scaled computational graph framework.

\paragraph{Glorot initialisation \citep{Glorot10}}
For a layer $Y = f(X W)$, consider the scales $\sigma_Y$ and $\sigma_{\nabla_X L}$, ignoring $\sigma_{\nabla_W L}$. Apply full constraints, and typically use arithmetic mean rather than geometric mean to combine scales. Finally, push the combined scale into the initialisation of $W$, so that no multiplication is required at execution time.

\paragraph{Loss scaling \citep{Micikevicius18}}
Introduce a single scaled identity op before the loss. $f^*(x) = \alpha \cdot x$, $f^*_{\mathrm{grad}}(x, g) = \beta \cdot g$. Since this edge is always a cut-edge, set $\alpha = 1$, and use $\beta$ to generate gradients that all share a single scale. Unlike unit scaling, there are no local distributional assumptions that can inform the choice of loss scale---it must be chosen empirically or heuristically.

\paragraph{Scaled dot product self attention \citep{Vaswani17}}
When computing the similarity matrix $A = Q K^{\top},\, Q, K \in \mathbb{R}^{s \times d}$, consider the scale $\sigma_A$, ignoring $\sigma_{\nabla_Q L}$, $\sigma_{\nabla_K L}$. Apply fully constrained scaling, yielding $\alpha = \beta_1 = \beta_2 = \frac{1}{\sqrt{d}}$. This is perhaps the best pre-existing example of a commonly employed scheme similar to unit scaling.

\section{Unit scaled ops compendium}
\label{app:ops_compendium}

Unit scaling relies on the correct selection of the scaling factors $\alpha, \beta_i, \dots, \beta_k$ for a given op. These scaling factors are derived from an analysis of the scaling of a given operation and its corresponding grad op, as outlined in Section~\ref{sec:scaling_strategy}, with an example of analysing the scaling of a multilayer perceptron given in Appendix~\ref{app:theory}.

To avoid practitioners having to analyse the scaling characteristics of each op in their model by hand, we provide a reference for common ops in Table~\ref{tab:ops_compendium}, giving scaled versions of each op alongside necessary scaling factors.

We provide further details on the derivation of certain non-trivial scaled operations below.

\paragraph{Activations}

We calculate the scaling of $\textrm{ReLU}$ analytically, based on the analysis in Appendix~\ref{app:scaling_example}. The other activation functions given are not amenable to the same procedure, so we calculate their scaling empirically (this is done through the use of short programs, which only need consider functions in isolation rather than within a larger model).

\paragraph{Softmax (followed by matmul)}

We make the simplifying assumption in our analysis that the output of a softmax over $s$ normally-distributed elements is uniformly $1/s$. In practice, there is some variance across output elements but this is small enough to ignore for our purposes.

This deviates from our standard unit scaling assumption of zero mean and unit variance, with $1/s$ mean and zero variance instead. Hence we require a different strategy for scaling softmax if we wish to still propagate unit scale.

We assume in this scenario that the softmax is followed by a matmul (as in multi-head self-attention). Based on this assumption, we scale by a factor of $s$, meaning the output is approximately a vector of ones.

From the perspective of the subsequent matmul, its ideal choice of scaling factor is then identical to the scaling factor it would have required if its input were sampled from a unit normal distribution: $m^{-\frac{1}{2}}$, where $m$ is the size of the dimension reduced over. The subsequent matmul op can then be implemented using our standard scaling without any special-case behaviour.

We also find through empirical analysis that the backward pass of softmax requires $s$ scaling, though in this direction it generates normally distributed values, conforming to our standard assumption.

\paragraph{Softmax cross-entropy}

We now consider a softmax going into a cross-entropy function, treating this composition as a single operation: $\mathrm{softmax\_xent}(x, t) = -\log \mathrm{softmax}(x)_t$ (where $t$ is the index of the target label), and assume that this is the final layer in a model used to generate a loss.

On this basis, we need not consider forward scaling, and focus on the backward operation $x^{\prime} = \textrm{softmax\_xent}_{\textrm{grad}}(x, t)$ and the calculation of its scaling factor $\beta = 1 / \sigma(x^{\prime})$. 

Assuming again that at the beginning of training the output of the softmax over $s$ inputs is uniformly $1/s$, the gradient of softmax cross-entropy is given by,
\[
x^{\prime} = \textrm{softmax\_xent}_{\textrm{grad}}(x, t)_i=
\begin{cases}
			\frac{1-s}{s}, & \text{if $i=t$}\\
            \frac{1}{s}, & \text{otherwise}
\end{cases}
\]
where $x \in \mathbb{R}^s$.

To calculate $\sigma(x^{\prime})$ we first observe that,
\begin{align*}
    \mathbb{E}\left[x^{\prime}\right] &= \frac{1}{s}\left(\frac{1-s}{s} + (s-1) \frac{1}{s}\right)\\
    &= 0
\end{align*}
from which we derive,
\begin{align*}
    \sigma(x^{\prime})^2 &= \mathbb{E}\left[(x^{\prime})^2\right] - \mathbb{E}\left[x^{\prime}\right]^2\\
    &= \frac{1}{s}\left(\left(\frac{1-s}{s}\right)^2 + (s-1) \left(\frac{1}{s}\right)^2\right) - 0\\
    &= \frac{1}{s}\left(\frac{1-2s+s^2 + s-1}{s^2}\right)\\
    &= \frac{s-1}{s^2}
\end{align*}

This gives us our scaling factor, $\beta = s/\sqrt{s-1}$.

\begin{table*}
\caption{Table of unit scaling factors, based on simple distributional assumptions on inputs and gradients, most often that they are unit normal.}
\label{tab:ops_compendium}
\centering
\vspace{0.6em}
\renewcommand{\arraystretch}{1.25}
\begin{tabular}{p{8cm}p{6cm}}
    \toprule
    Op & Unit scaling factors \\
    \midrule
    
    {\hspace{0.5cm}\small\textsc{Linear}}\\
    
    $\mathrm{matmul}(X^{b \times m}, W^{m \times n})^{b \times n} = X W$
    & $\alpha=m^{-\frac{1}{2}}$, $\beta_X=n^{-\frac{1}{2}}$, $\beta_W=b^{-\frac{1}{2}}$ \\
    
    $\mathrm{sum}(x) = \sum_{i=1}^n x_i$
    & $\alpha=n^{-\frac{1}{2}}$, $\beta=1$ \\
    
    $\mathrm{weighted\_add}(x_{i\in1\dots n}, \gamma_{i\in1\dots n}) = \sum_{i=1}^n \gamma_i x_i$
    & $\alpha=\left(\sum_i \gamma_i^2\right)^{-\frac{1}{2}}$,
    $\beta_i=\gamma_i^{-1}$ \\
    
    \midrule
    {\hspace{0.5cm}\small\textsc{Activations}}\\
    
    $\mathrm{relu}(x) = \mathrm{max(x, 0)}$
    & $\alpha=\sqrt{2 /\left(1-1/\pi\right)}$, $\beta=\sqrt{2}$ \\
    
    $\mathrm{gelu}(x) = x \cdot \Phi(x)$
    & $\alpha=1.701$, $\beta=1.481$ \\
    
    $\mathrm{tanh}(x) = (e^{2 x} - 1)/(e^{2 x} + 1)$
    & $\alpha=1.593$, $\beta=1.467$ \\
    
    $\mathrm{sigmoid}(x) = (1 + e^{-x})^{-1}$
    & $\alpha=4.802$, $\beta=4.722$ \\
    
    \midrule
    {\hspace{0.5cm}\small\textsc{Other}}\\
    
    $\mathrm{softmax}(x)_i = e^{x_i}/\sum_{j=1}^s e^x_j$
    & $\alpha=s$, $\beta=s$\\
    
    $\mathrm{softmax\_xent}(x, t) = \log \mathrm{softmax}(x)_t$
    & $\alpha=1$, $\beta=s / \sqrt{s-1}$\\
    
    $\mathrm{layer\_norm}(X^{b \times n}, w, c)_{ij} = c_j + w_j \cdot (X_{ij} - \mu_i) / \sigma_i$,\newline
        \ldots $\mu_i = \frac{1}{n}\sum_{j=1}^n X_{ij}$,\,
        $\sigma_i = \sqrt{\frac{1}{n}\sum_{j=1}^n X_{ij}^2 - \mu_i^2}$
    & $\alpha = 1$, $\beta_x = 1$, $\beta_w = \beta_c = b^{-\frac{1}{2}}$ \\
    
    \bottomrule
\end{tabular}
\end{table*}

\section{Aligning unit scaling with existing models}
\label{app:aligning}

Our presentation of unit scaling in Section~\ref{sec:unit_scaling} assumes the design of a model from scratch. However, we anticipate there will be cases in which practitioners will wish to unit scale existing models, such that their unit scaled model and base model are either equivalent or similar enough to give matching performance.

Here we outline the additional considerations required to do so. We follow this approach for our BERT experiments in Section~\ref{sec:mlm}.

\subsection{Activation functions}

We take `activation function' to mean any non-linear element-wise function in a model.
Due to non-linearity, the behaviour of an activation function $f(x)$ depends on the scale of its input. Therefore a base model's activation functions may not have the same effect on their inputs as a unit scaled version, as the unit scaled model alters the scale of inputs.

To counter this, one can introduce a scaling factor immediately before an activation function (temporarily breaking unit scale), and a second un-scaling factor immediately afterwards (restoring unit scale):
\[
\hat{f}(\hat{x}) = f(s_1 \cdot \hat{x}) \cdot s_2,
\]
where $\hat{f}$ is our new `aligned' activation function, $\hat{x}$ is assumed to be normally distributed with unit scale (not necessarily true for $x$ in the base model), and $s_1, s_2 \in \mathbb{R}$ are our new scaling factors.

We select the first scaling factor such that $s_1 = \sigma(x)$, giving identical-scale inputs to both activation functions: $\sigma(s_1 \cdot \hat{x}) = \sigma(x)$.

The second scaling factor is selected to restore unit scale: $s_2 = \frac{1}{\sigma(f(x))}$, giving,
\begin{align*}
    \sigma(\hat{f}(\hat{x})) &= \frac{f(\sigma(x) \cdot \hat{x})}{\sigma(f(x))},\\
    &= 1.
\end{align*}

All that remains is the estimation of $\sigma(x)$ and $\sigma(f(x))$ in the base model. This can be done either analytically (by stepping through operations in the base model and calculating the expected scale at each) or empirically (via instrumentation of the base model). The latter method tends to be simpler and less error-prone, but the former is more mathematically rigorous and has the advantage of generating scaling factors that are a function of the model's hyperparameters.

Note that although we temporarily break the assumption of unit scale in the above analysis, in practice scaling factors here are close enough to 1 that this momentary mis-scaling is negligible from a numerics perspective.

\subsection{Softmax functions}

The above analysis also applies to softmax functions. Although softmax is not an element-wise function, the same approach is still valid and $s_1, s_2$ should be chosen in the same way.

Note that the standard softmax function is sometimes introduced with a `temperature' scalar $T$, by which all inputs are divided. Hence our method can be seen as tuning the effective temperature of the softmax to align the unit scaled model with the base model.

\subsection{Residual weighted add}

In Section~\ref{sec:weighted_addition} we recommended that practitioners introduce a weighted addition into their models between residual and skip branches, in order to actively select how much each contributes to the output.

A typical unscaled base model implicitly makes this choice via the scaling effect of the residual branch (i.e. the ratio of $\sigma(f(x)) / \sigma(x)$, which typically $\neq 1$).

For our unit-scaled model to be equivalent to the base model, we need the output of our addition to be equal up to a constant (unit) scaling factor $\alpha$.

Taking a \textit{fixed($\tau$)} residual layer, this means we must maintain: $\sqrt{1-\tau} \, \hat{x} + \sqrt{\tau} \, \hat{f}(\hat{x}) = \alpha (x + f(x))$, where $\hat{f}(\cdot)$ is the residual branch and $\hat{x}$ the input in our unit-scaled model.

Thanks to unit scaling, we have $\hat{x} = x/\sigma(x)$ and $\hat{f}\hat{(x)} = f(x) / \sigma(f(x))$ giving,
\[
\sqrt{1-\tau} \, \hat{x} + \tau \, \hat{f}(\hat{x}) = \sqrt{1-\tau} \, \frac{x}{\sigma(x)} + \sqrt{\tau} \, \frac{f(x)}{\sigma(f(x))}
\]
Our desired form requires the terms multiplying $x$ and $f(x)$ to be equal, meaning:
\begin{align*}
    \frac{\sqrt{1-\tau}}{\sigma(x)} &= \frac{\sqrt{\tau}}{\sigma(f(x))}\\
    \tau &= \frac{\sigma(f(x))^2}{\sigma(x)^2 + \sigma(f(x))^2},\\
\intertext{giving,}
    \alpha &= \frac{1}{\sqrt{\sigma(x)^2 + \sigma(f(x))^2}},\\
\end{align*}
and recalling that our original definition of a \textit{fixed($\tau$)} residual layer ensures that this still maintains a unit-scaled output.

Hence to align the residual add operation with a base model, we need first need to use a \textit{fixed($\tau$)} residual layer, and secondly calculate $\sigma(x)$ and $\sigma(f(x))$ for the base model, plugging them into the above equation for $\tau$.

This calculation of $\sigma$ in the base model can again be done analytically or empirically. For typical models, the correct value of $\tau$ is the same across layers.

\subsection{Shared parameters}

Weights used in multiple operations in the forward pass sum the weight gradients coming from those operations in the backward pass.

The same argument used for the residual add applies to the alignment of this summation too: for a unit-scaled model to be equivalent it must match the ratio of scales going into this sum as in the base model. Unit scaling will normalise these all to have $\sigma = 1$, but this is not guaranteed in the base model.

The same analysis as used for the residual add op can be applied here, with the same outcome. The calculation of the scale of residual branches in the base model should be substituted with the scale of each weight gradient. In the case that the weight gradient is used more than twice, the above argument will have to be generalised to multiple operands.

\subsection{Example: aligning BERT}
\label{app:example_align_BERT}

We follow the steps above in our experiments for Section~\ref{sec:mlm}, where we align unit-scaled BERT models against standard baseline models, to match performance.

Here we outline how we apply the above rules in practice, along with a few additional considerations required due to specifics relating to the BERT architecture.

Where these rules require the calculation of standard deviation of tensors in the base model, we always calculate them analytically, rather than relying on empirical measurements (though we have then used empirical measurements to check the correctness of our calculations).

\paragraph{Embedding layer}

BERT contains three separate embeddings: a general word embedding, along with segment and positional embeddings. These are all combined using a summation at the beginning of the model. For unit scaling we must implement this using:

\[
x_{\textrm{emb}} = \mathrm{weighted\_add}\left(x_{\textrm{word}}, x_{\textrm{seg}}x_{\textrm{pos}}, \frac{1}{\sqrt{3}}, \frac{1}{\sqrt{3}}, \frac{1}{\sqrt{3}}\right).
\]

Weights are equal here as the initial scales of the embeddings in the base model are unchanged from their initialisation, and all are initialised with the same scale.

\paragraph{FFN}

For the FFN, alignment need not be considered for the matmul and layernorm ops, which we scale using the set of scaling factors for common ops given in Table~\ref{tab:ops_compendium}. For the gelu activation function, we must follow the alignment process outlined above, applying scaling factors immediately before and after.

\paragraph{Multi-head self-attention}

For multi-head self attention, we employ the rule for aligning softmax (followed by a matmul) given above. Again, matmuls do not require alignment with the base model. We note that in the particular case of the matmul with the $V$ tensor, our standard distributional assumption of independent elements no longer strictly holds, due to correlation across the sequence dimension introduced by the segment embedding. This requires a slight correction to ensure unit scaling is maintained.

\paragraph{Residual connection}

Both the FFN and multi-head self-attention layers are residuals, and as such employ the rule above for aligning weighted addition with a base model.

\paragraph{Loss heads}

We train BERT according to the standard procedure of using two heads: one for the masked-language-modelling (MLM) task, and one for the next-sentence-prediction (NSP) task. The NSP head uses a tanh activation function which requires alignment, and the MLM head re-uses the weights of the word embedding for a matmul, requiring the above rule for aligning shared parameters. Each head is terminated by a softmax cross-entropy, that we also tune to match the base model.

\paragraph{Sequence length considerations}

Care must be taken when unit-scaling sequence-based models to account for the role of the sequence dimension. For many ops this effectively becomes an extra batch dimension, and must be handled as such when applying unit scaling.

In our experiments we use padding to compensate for uneven-length input-sequences. In this case the value used for our sequence calculations is not the length of the sequence dimension, but the average number of non-padding tokens in a sequence (for our experiments, this was approximately $77\%$ of the padded length).

One additional complication specific to BERT, is that the gradients flowing back into the final transformer layer are sparse, as they only come via the subset of tokens used in the two heads (specifically, the \texttt{[CLASS]} token, and those tokens masked for the MLM head). As a result, backwards-pass sequence length calculations for this layer must be adapted to assume a smaller sequence length, according to the level of sparsity in the gradient.

\section{Implementation}
\label{app:code}

Unit scaling is straightforward to implement in deep learning frameworks such as PyTorch, JAX and TensorFlow, that support user-defined custom gradient autograd operations. A convenient way to do this is via a scaled identity op $\mathrm{id}^*(x, \alpha, \beta)$, which can be used to implement scaled ops without defining custom gradients for each.

\subsection{Code examples}
We show an example implementations in Figure~\ref{code:main}, with additional code listings in Figure~\ref{code:appendix}, demonstrating basic tools for constructing unit-scaled models in PyTorch. Note:

\texttt{scaled} is the basic building block of unit-scaled models. It enables independent control of forward and backward pass scaling factors, and as such must be used with care---it could be used to define a scaled graph with incorrect constraints, leading to gradients that are inconsistent with the forward pass of the model.

\texttt{scaled\_matmul} demonstrates how to combine multiple constraints using geometric mean.

\texttt{scaled\_gelu} implements only fully constrained scaling, for brevity. When scales are fully constrained, custom gradients via \texttt{scaled} are optional. Note that it may still be useful in certain situations for improving the scale of intermediate values.

\texttt{ScaledLayerNorm} uses the usual assumption for scaled layers: weights are cut-edges, activations are not. This permits independent scales for the weight and bias parameters.

\begin{figure}[tbp]
\codefig{appendix.py}
\caption{Definition of \texttt{scaled} in PyTorch, as a custom autograd function. Additional scaled ops and layers required for a Transformer FFN. See Table~\ref{tab:ops_compendium} for a reference of scaling factors.}
\label{code:appendix}
\end{figure}

\subsection{Computational overhead}
\label{app:overhead}

Unit scaling typically introduces one extra function invocation per invocation in the equivalent unscaled model. For example, matmul typically involves 3 function invocations during training, corresponding to $1\times$ forward, $2\times$ backward functions (one for each input). Using unit scaling, there are $3$ additional \textit{rescaling} function invocations of the form $f(x, \gamma) = \gamma \cdot x$, where $\gamma \in \mathbb{R}$, $x \in \mathbb{R}^n$.

\paragraph{FLOPs}
Considering the typical theoretical metric for computational effort, \textit{floating point operations} (FLOPs), the overhead appears much smaller. For the matmul op with forward pass $matmul : \mathbb{R}^{b \times n} \times \mathbb{R}^{n \times m} \to \mathbb{R}^{b \times m}$, the amount of computational effort due to $3\times$ matmul is $6 \, b \, n\, m$ (note this is $2\times$ because multiply and add are counted separately), while rescaling consumes $b n + n m + b m$. Therefore the ratio of rescaling to matmul flops follows:
\begin{equation*}
    \frac{\mathrm{FLOP_{rescaling}}}{\mathrm{FLOP_{matmul}}} = \frac{1}{6} (b^{-1} + m^{-1} + n^{-1}).
\end{equation*}
Note that this is bounded above by $(2\cdot\mathrm{min}(b, n, m))^{-1}$. For the matmuls that dominate compute in many models, this minimum dimension corresponds to the hidden size.

There are also operations other than matmuls that require scaling, but contribute negligible FLOPs. To simplify analysis, we'll assume that there are $(\mathrm{ops\_per\_matmul} - 1)$ additional ops for every matmul in the model.

So we write $\mathrm{FLOP_{matmul+}} \approx \mathrm{FLOP_{matmul}}$ and $\mathrm{FLOP_{rescaling+}} = \mathrm{ops\_per\_matmul} \cdot \mathrm{FLOP_{rescaling}}$. This gives the following adjusted estimate for the FLOP overhead of unit scaling a model:
\begin{equation*}
    \frac{\mathrm{FLOP_{rescaling}}}{\mathrm{FLOP_{unscaled}}} = \frac{\mathrm{ops\_per\_matmul}}{2 \cdot \mathrm{hidden\_size}}.
\end{equation*}
In the example of BERT\textsubscript{LARGE}, we set $\mathrm{hidden\_size} = 1024$, pessimistically estimate $\mathrm{ops\_per\_matmul} = 4$, and obtain a FLOP overhead of 0.2\%.

Other large models should behave in a similar manner, so we conclude that the theoretical FLOP overhead of unit scaling is small for large models. Actual performance will depend on many other factors, and we anticipate that FLOP-based measures are likely to be optimistic in predicting runtime overhead on typical deep learning hardware. However, we expect the efficiency gains of low-precision formats to vastly outweigh the scaling overhead.

\paragraph{Fusing scale factors}
We anticipate substantial efficiency gains from fusing the fixed scale factors from unit scaling into preceding ops. This yields two potential benefits. First, fusing avoids the communication overhead of an extra round-trip to memory. Second, it may permit low-precision outputs and even intermediate values. This may be particularly valuable for distributed aggregation ops, where partial results are aggregated on separate workers before sharing them to compute a final result.

Transformations implementing automatic fusing of ops are widely available using optimising compilers such as XLA \citep{Xla17}. These are particularly effective at fusing consecutive elementwise ops, which should encompass most unit scaling factors (since \texttt{matmul} outputs are typically first used in \texttt{add} or activation functions).

\section{Additional experimental details and results}
\subsection{Character language modelling}
\label{app:char}

The WikiText-103 raw dataset consists of approximately 500 million characters of text extracted from Wikipedia articles. We do not perform any additional preprocessing beyond that of the published dataset. All results correspond to the best value over a learning rate sweep starting from a low value, with step $\times 2$. A complete set of hyperparameters used is shown in Table~\ref{tab:char_hyperparameters}.

\paragraph{Mixed precision} When running in FP16, all activations, parameters and gradients are stored in FP16. Optimiser state is also stored in FP16, with the exception of Adam's second moment state, which is stored in FP32 since squared values are more prone to clipping.

\paragraph{Model architectures} All models are based on causal Transformer-like stacks that interleave contextual (i.e. token-mixing) layers and FFN layers. Input tokens are embedded by indexing into a trainable embedding table, and output token probabilities are generated by $\mathrm{softmax}(W_{\mathrm{proj}}\, \mathrm{layernorm}(x_L) + b_{\mathrm{proj}})$, where $x_L$ is the final hidden state from the Transformer stack.

The basic unscaled layer definition follows:
\begin{align*}
    x_{l+1} &= \mathrm{res}(\mathrm{ffn}, \mathrm{res}(\mathrm{context}, x_l)) \\
    \mathrm{res^{NoNorm}}(f, z) &= \mathrm{interp}(z, f(z)) \\
    \mathrm{res^{PreNorm}}(f, z) &= \mathrm{interp}(z, f(\mathrm{layernorm}(z))) \\
    \mathrm{res^{PostNorm}}(f, z) &= \mathrm{layernorm}(\mathrm{interp}(z, f(z))) \\
    \mathrm{interp^{default}}(a, b) &= a + b \\
    \mathrm{interp^{fixed}}(a, b; \tau) &= \sqrt{1-\tau} \cdot a + \sqrt{\tau} \cdot b \\
    \mathrm{interp^{mean}}(a, b; l) &= \sqrt{l/(l+1)} \cdot a + \sqrt{1/(l+1)} \cdot b \\
    \mathrm{ffn}(z) &= W_2\, \mathrm{max}(0, W_1\, z + b_1) + b_2
\end{align*}

The contextual layers are as follows:

\begin{enumerate}
    \item $\mathrm{context^{Attention}}$: multi-head dot product self attention using causal masking \citep{Vaswani17}, with relative-positional encoding using sinusiodal bases following \citet{Dai19},
    \item $\mathrm{context^{Conv}}$: 1D grouped causal convolution with relu nonlinearity,
    \item $\mathrm{context^{RNN}}$: recurrent highway network \citep{Zilly17} with tied transform and carry gates $x_{t+1} = (1 - g(x_t)) \cdot x_t + g(x_t) \cdot f(x_t)$, where $g(x)$ is a projection with sigmoid nonlinearity, and $f(x)$ is a projection with tanh nonlinearity.
\end{enumerate}

When applying unit scaling, we also reduce the learning rate for non-projection parameters by $1/\sqrt{\mathrm{hidden\_size}}$ to compensate for the relative step size increase implied by unit scaling.

\paragraph{Additional results}
Test set results, with multiple runs per learning rate are shown in Table~\ref{tab:char_results_test}. These support the main findings shown for the wider sweep of Figure~\ref{fig:char_sweep}: unit-scaled models perform comparably to regular models, and can be trained in FP16 without modification or additional hyperparameter selection.

\begin{figure}[tbp]
    \centering
    \includegraphics[width=8.2cm]{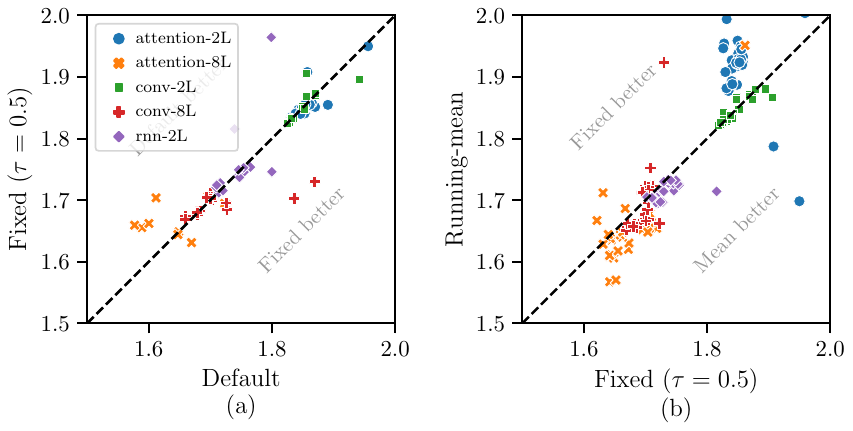}
    \caption{Comparison of residual scaling approaches. We observe (a) for regular models, default scaling performs similarly to fixed interpolation $\tau=0.5$; (b) in most cases, running-mean scaling is similar or better than fixed interpolation. The exception is 2-layer attention models, where we hypothesise that running mean places too much weight on the first layer, which is detrimental in such a shallow model.}
    \label{fig:char_residual}
\end{figure}

Figure~\ref{fig:char_residual} shows the effect of employing residual scaling schemes described in Section~\ref{sec:scaling_strategy}. This supports the claim that fixed and running-mean residual scaling are viable alternatives to default scaling, since both perform well in regular and unit-scaled models.

\begin{table}[tbp]
    \caption{Character language modelling hyperparameters.}
    \label{tab:char_hyperparameters}
    \vspace{1em}
    \centering
    \begin{tabular}{p{4cm}p{3.2cm}}\toprule
        Parameter & Value \\
        \midrule
        Sequence length & 256 characters \\
        Sequence mask & 32 characters \\
        Batch size & 2048 characters \\
        Training duration & $2^{19}$ steps \\
        Learning rate decay half-life & $2^{16}$ steps \\
        Adam $(\beta_1, \beta_2)$ & (0.9, 0.999) \\
        SGD momentum & 0.9 \\
        \midrule
        Vocabulary size & 5008 characters (100\% coverage, no OOV) \\
        Hidden size & 128 \\
        FFN size & 512 \\
        Depth & [2, 8] layers \\
        \midrule
        Attention heads & 2 \\
        Attention head size & 64 \\
        Relative positional frequency components & 128 bases, period [1 \dots 1024] characters \\
        Convolution kernel size & 7 \\
        Convolution group size & 16 \\
        \midrule
        Typical learning rate ranges: & \\
        Regular, SGD & $2^{-8} \dots 2^{-4}$ \\
        Regular, Adam & $2^{-12} \dots 2^{-8}$ \\
        Unit, SGD & $2^{-14} \dots 2^{-10}$ \\
        Unit, Adam & $2^{-8} \dots 2^{-4}$ \\
        \bottomrule
    \end{tabular}
\end{table}

\begin{table}[tbp]
    \caption{Character language modelling, test BPC with 3 runs per learning rate. The best learning rate is chosen according to validation BPC. 95\% confidence interval is $\pm 0.010$. All models use PreNorm and 8 layers, except where noted.}
    \label{tab:char_results_test}
    \vspace{1em}
    \centering
    \begin{tabular}{p{3cm}p{1.2cm}p{1.2cm}p{1.2cm}}\toprule
        Model & Regular FP32 & Unit scaling FP32 & Unit scaling FP16 \\
        \midrule
        Attention (PostNorm) & 1.548 & 1.540 & 1.540 \\
        Attention & 1.582 & 1.562 & 1.567 \\
        Convolution & 1.625 & 1.620 & 1.622 \\
        RNN (2 layers) & 1.674 & 1.677 & 1.673 \\
        \bottomrule
    \end{tabular}
\end{table}

\begin{table}[tbp]
    \caption{BERT pre-training hyperparameters.}
    \label{tab:mask_hyperparameters}
    \vspace{1em}
    \centering
\begin{tabular}{p{3cm}p{4.2cm}}
\toprule
Parameter & Value \\
\midrule
Sequence length             &                    [128, 384] tokens \hfill (phase 1/2) \\
Depth                       &                            [12, 24] \hfill (base/large) \\
Hidden size                 &                         [768, 1024] \hfill (base/large) \\
FFN size           &                        [3072, 4096] \hfill (base/large) \\
Attention heads             &                            [12, 16] \hfill (base/large) \\
Attention head size         &                                                 64 \\
Vocabulary size             &                                              30400 \\
\midrule
Total batch size            &                        [16320, 4080] seqs \hfill (ph. 1/2) \\
Micro-batch size            &                               [8, 2] \hfill (phase 1/2) \\
Data-parallel count         &                                                  4 \\
Gradient accumulation count &                                                510 \\
\midrule
Training duration           &                  [28266, 8437] steps \hfill (ph. 1/2) \\
Learning rate               &                     [0.0045, 0.0015] \hfill (phase 1/2) \\
Warmup steps  &                    [2827, 275] steps \hfill (phase 1/2) \\
Learning rate decay  &                                             linear \\
\midrule
Optimiser                   &                                               LAMB \\
LAMB Beta1                  &                                                0.9 \\
LAMB Beta2                  &                                              0.999 \\
LAMB epsilon                &                                              1e-06 \\
Weight decay                &                                               0.01 \\
\midrule
Weight init std             &                          0.02 \hfill (unit scaling=n/a) \\
Loss scaling                &  [512, 512, 32768, 128] \;\;\; (base phase 1/2, large phase 1/2; \; unit scaling=n/a) \\
\bottomrule
\end{tabular}
\end{table}

\subsection{Masked language modelling}
\label{app:mask}

We follow the standard practice of splitting BERT pre-training into two phases. For the first phase we use a sequence length of 128 tokens, and for the second we use 384. Tokens are derived using the WordPiece tokeniser \citep{Wu16}, with a vocabulary of 30400 tokens. Our masking approach is consistent with that used in \citet{Devlin19}. A complete set of pretraining hyperparameters used is shown in Table~\ref{tab:mask_hyperparameters}.

\paragraph{Mixed precision} For FP16, we follow the same approach here as in our character language modelling experiments (appendix \ref{app:char}), storing all tensors and optimiser state in FP16, apart from the optimiser second moment state which is stored in FP32 (note, we use the LAMB optimiser \citep{You2019} here over Adam).

For FP8, we modify our FP16 mixed precision strategy by quantising the inputs to all matmul operations. Note that our experiments do not utilise hardware FP8 support; we instead simulate FP8 training by quantising from FP16 to the set of supported values in a given FP8 format. In this, we are following the approach taken by \citet{Noune22} and \citet{Micikevicius22}. As recommended in both studies, we also use E4 for activations and weights, and E5 for all gradients. Again, following the precedent set in these studies, the one matmul operation we exclude from FP8 quantisation is the vocabulary embedding matmul, which has been known to cause numerical instabilities.

\paragraph{Hardware \& distributed training}

Models were trained on IPU hardware \citep{Jia2019}, using either Bow~Pod\textsubscript{16} or IPU-POD16 Classic machines. On each machine we distribute training across 16 IPUs, using 4-way model parallelism and 4-way pipeline parallelism, with gradient accumulation across pipeline stages.

\section{Histograms of tensor-scaling within BERT}
\label{app:bert_scaling}

To give readers a better intuitive sense of how loss scaling and unit scaling operate for a standard model, we provide histograms of absolute tensor values taken from FP16 BERT\textsubscript{BASE}.

Figures~\ref{fig:bert_scaling_reg_init} and~\ref{fig:bert_scaling_us_init} show the beginning of training for loss and unit scaling respectively, and Figures~\ref{fig:bert_scaling_reg_end} and~\ref{fig:bert_scaling_us_end} show the end of training.

We use 9 transformer layers rather than the standard 12 in order to accommodate the overheads of tracking histograms across all tensors in the model. For the sake of concision we omit histograms of the middle layers, which are substantially similar to layers 0 and 7 in both the forward and backward pass. A small number of numerically insignificant ops are also omitted.

The first two figures can be understood as the full-model equivalent to the plot in Figure~\ref{fig:illustration}, with the second two showing how values shift as a result of training. The x-axis is labelled slightly differently to Figure~\ref{fig:illustration}, showing the log of absolute values rather than the exponent value, but by the definition of floating point values given in Section~\ref{sec:formats}, these two are approximately equivalent. We also have a special bin for the range $\left[2^{-24}, 2^{-14}\right]$, which represents all subnormal values in the FP16 range, and bins on either end to hold zero and infinity values.

There are some surprising features in the shapes of these plots, resulting from the design of BERT. We provide a brief analysis here of our key plot: Figure~\ref{fig:bert_scaling_us_init} (unit scaling at initialisation).

\begin{figure*}
\centering
\includegraphics[width=1.07\textwidth, trim = 6em 7em 0em 5em, clip]{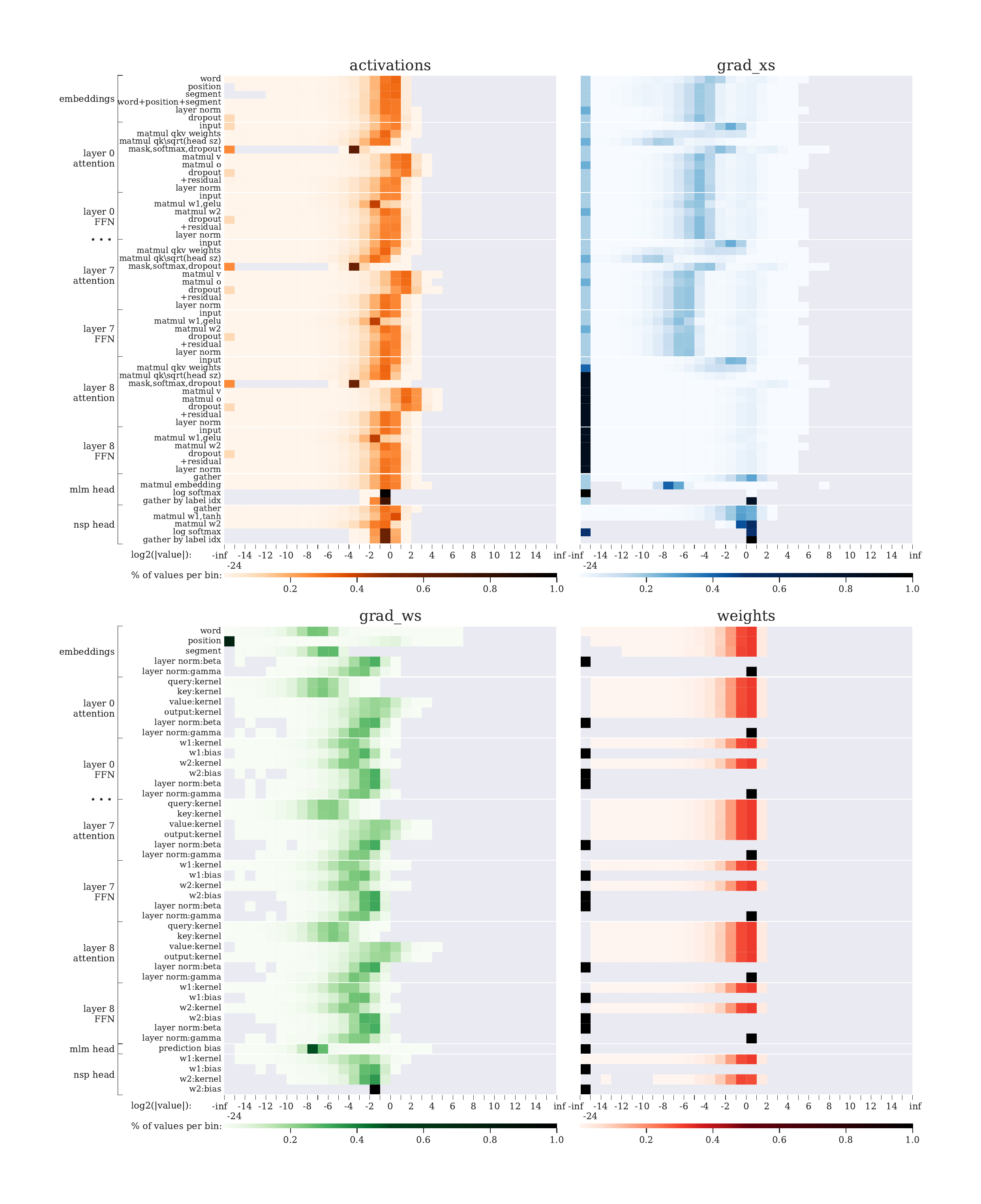}
\caption{A histogram of absolute values in \textbf{unit-scaled} BERT\textsubscript{BASE} at \textbf{initialisation}. Unit scaling naturally places values in approximately the centre of the range without requiring a tuned hyperparameter. See Appendix~\ref{app:bert_scaling_unit_analysis} for specific details of this plot.}
\label{fig:bert_scaling_us_init}
\end{figure*}

\begin{figure*}
\centering
\includegraphics[width=1.07\textwidth, trim = 6em 7em 0em 5em, clip]{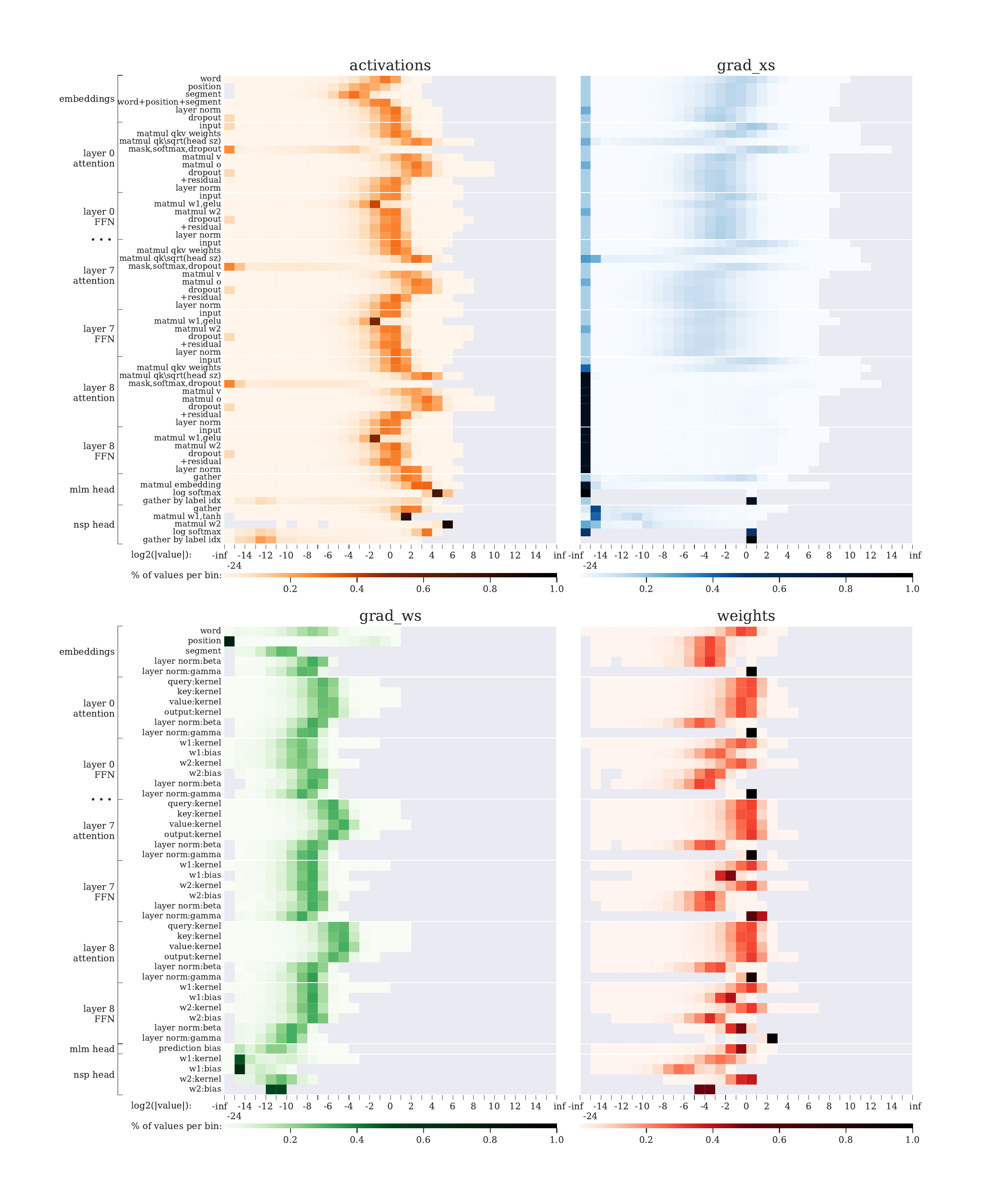}
\caption{A histogram of absolute values in \textbf{unit-scaled} BERT\textsubscript{BASE} at the \textbf{end of training}. Compare with figure \ref{fig:bert_scaling_us_init} to see the shift in distributions during training and the implications for numerics.}
\label{fig:bert_scaling_us_end}
\end{figure*}

\subsection{Analysis of Figure~\ref{fig:bert_scaling_us_init}}
\label{app:bert_scaling_unit_analysis}

\begin{figure*}[tb]
\centering
\includegraphics[width=1.07\textwidth, trim = 6em 7em 0em 5em, clip]{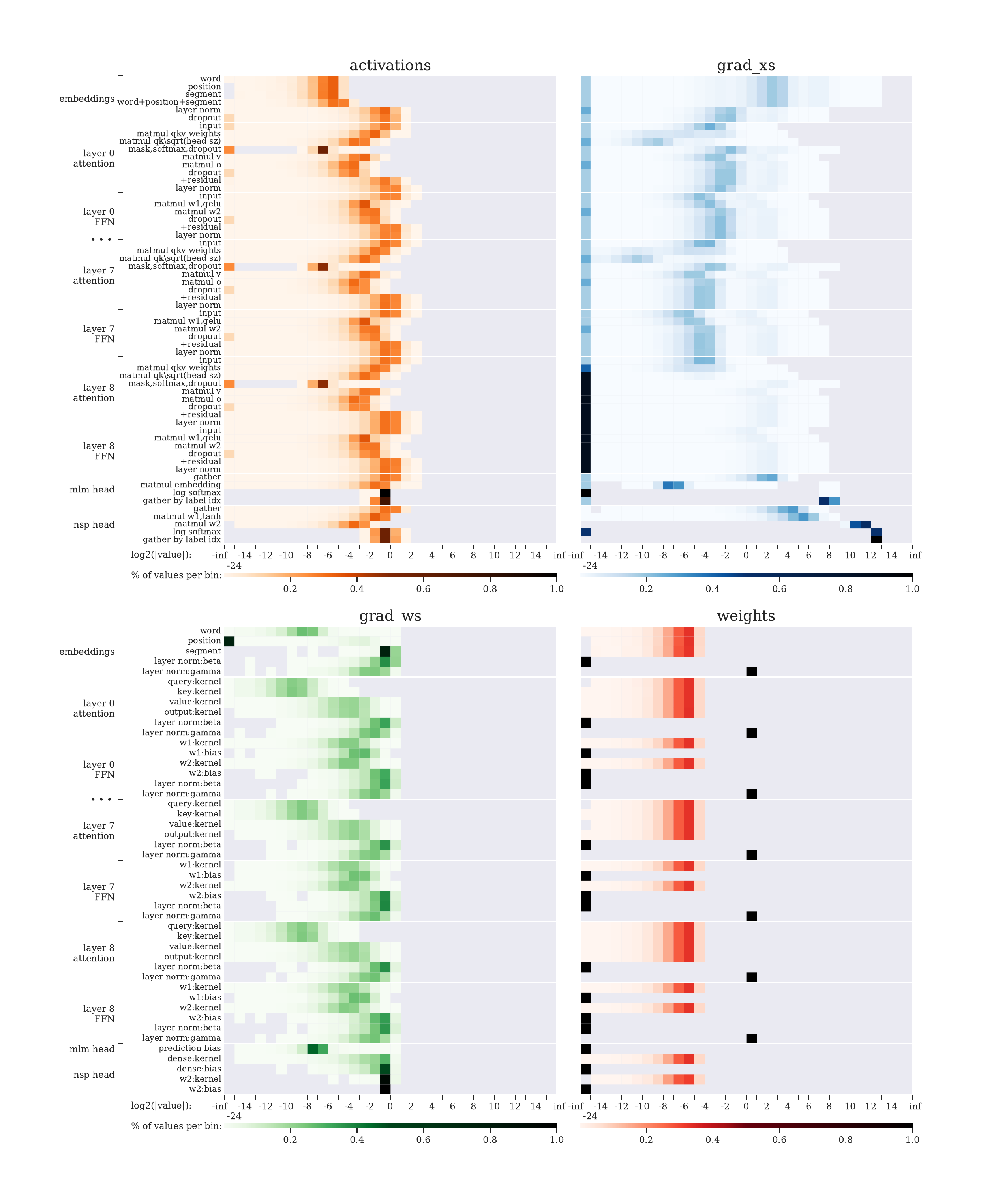}
\caption{A histogram of absolute values in \textbf{regular} BERT\textsubscript{BASE} at \textbf{initialisation}. Here a loss scale of $2^{15}$ was required for stable training. We can understand loss scaling in light of this plot as enacting a shift of the \textit{gradx} and \textit{gradw} histograms by $\log_2(\textrm{loss scale})$ to the right.}
\label{fig:bert_scaling_reg_init}
\end{figure*}

\begin{figure*}
\centering
\includegraphics[width=1.07\textwidth, trim = 6em 7em 0em 5em, clip]{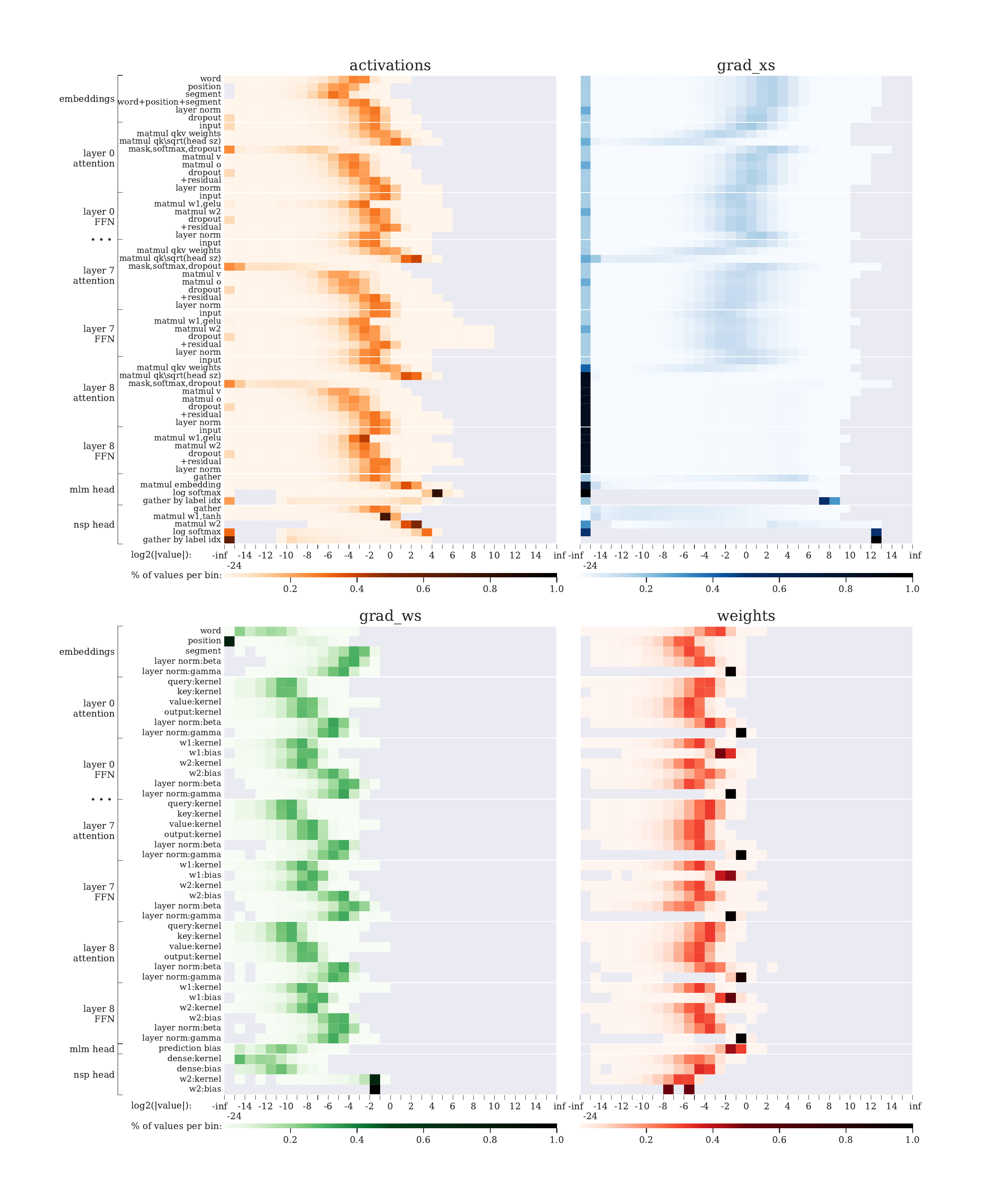}
\caption{A histogram of absolute values in \textbf{regular} BERT\textsubscript{BASE} at the \textbf{end of training}. Compare with figure \ref{fig:bert_scaling_reg_init} to see the shift in distributions during training and the implications for numerics.}
\label{fig:bert_scaling_reg_end}
\end{figure*}

\paragraph{Impact of unit scaling} A comparison with Figure~\ref{fig:bert_scaling_reg_init} demonstrates the effectiveness of unit scaling. Whereas the loss-scaled model has to tune a hyperparameter to centre the two gradient sub-plots, unit scaling does this naturally. Furthermore, values in the unit-scaled model are typically closer to the centre of the range. Loss scaling also has the problem of very large \textit{gradx} values in its NSP and MLM heads.

\paragraph{Effect of aligning with regular BERT}
As outlined in Appendix~\ref{app:example_align_BERT}, we take a range of measures to align our unit scaled model more closely with the regular BERT base model, so that their performance is similar. This has the impact of temporarily mis-scaling certain operations. This can be seen most clearly in the case of gelu, which requires a scaling factor for alignment, but as a result is slightly below unit-scale in the diagram.

\paragraph{Sparse gradients for layer 8}

The \textit{gradx} values for layer 8 in all plots have most of their values set to zero. This is a consequence of sparse gradients flowing back into this layer from the NSP and MLM heads, as described in Appendix~\ref{app:example_align_BERT}. The cross-sequence mixing of gradients in the multi-head self-attention layer has the effect of removing this sparsity, giving a strong signal for all subsequent layers.

\paragraph{Three groups of gradient scales}

Our final observation is somewhat subtle, but key to understanding both the shape of the \textit{gradx} plots, and the particular difficulties encountered when training BERT in low-precision.

We note that in the \textit{gradx} plots there are in effect three separate `columns' visible: a strong signal (i.e. many values) on the left, a faint signal through the centre, and a very small number of values on the right. This is a consequence of BERT's design, rather than of any scaling technique.

The right-hand column is a result of the natural up-scaling of gradients flowing from BERT's NSP head. BERT naturally has larger gradients flowing out of this head. Note that these gradients are sparse, representing only a single token-gradient in each sequence, but the signal is kept alive throughout the layers by the residual connection, resulting in this feature of the plot.

The central column comes out of the MLM head in a similar fashion. This is still sparse, but contains more token-gradients and hence gives a stronger signal. Finally the main left-hand column results from the mixing of gradients in the multi-head self-attention layer. This removes sparsity in the tensor, giving a stronger signal. However, the attention mechanism in BERT naturally lowers the scale of values, meaning this third signal is shifted to the left.

The existence of these three groups of gradients creates a trimodal distribution of exponent values. As most values are still concentrated in the left-hand column, our assumption of a single normal distribution is still sufficient, but we effectively have to balance the positions of these three columns, meaning that the backward pass does not fall into a single, neat column.

\end{document}